\documentclass[10pt,twocolumn, letterpaper]{article}
\usepackage{cvpr}
\usepackage{times}
\usepackage{epsfig}
\usepackage{graphicx}
\usepackage{amsmath}
\usepackage{amssymb}
\usepackage{mathtools}
\usepackage{breqn}
\usepackage[usenames]{color}
\usepackage{algorithm}
\usepackage{algorithmic}
\usepackage{epstopdf}
\usepackage{subcaption}
\usepackage{multirow}
\usepackage[table, dvipsnames]{xcolor}
\usepackage[title, titletoc]{appendix}
\usepackage{arydshln}
\DeclareMathOperator*{\argmin}{argmin} 

\usepackage[pagebackref=true,breaklinks=true,letterpaper=true,colorlinks,bookmarks=false]{hyperref}

\cvprfinalcopy % *** Uncomment this line for the final submission

\def\cvprPaperID{0067} % *** Enter the CVPR Paper ID here

\ifcvprfinal\fi
\begin{document}

%%%%%%%%% TITLE
\title{Template Matching with Deformable Diversity Similarity}

\author{Itamar Talmi\thanks{Authors contributed equally}\\
Technion\\
{\tt\small titamar@campus.technion.ac.il}
% \and
% Roey Mechrez\footnotemark[1]\\
% Technion\\
% {\tt\small roey@tx.technion.ac.il}
% \and
% Lihi Zelnik-Manor\\
% Technion\\
% {\tt\small lihi@ee.technion.ac.il}
% }
\and
Roey Mechrez\footnotemark[1]\\
Technion\\
{\tt\small roey@tx.technion.ac.il}
\and
Lihi Zelnik-Manor\\
Technion\\
{\tt\small lihi@ee.technion.ac.il}
}

\maketitle
%\thispagestyle{empty}

%\newcommand*{\argmin}{arg\,min}

%==============================================================
%%% comment \ShowNotes out to remove all colored comments defined with \newcommand below %%%
\newcommand*{\ShowNotes}{}
%=================================================================

% maybe requires \usepackage[usenames]{color}
\definecolor{darkred}{rgb}{0.7,0.1,0.1}
\definecolor{darkgreen}{rgb}{0.1,0.7,0.1}
\definecolor{cyan}{rgb}{0.7,0.0,0.7}
\definecolor{dblue}{rgb}{0.2,0.2,0.8}
\definecolor{maroon}{rgb}{0.76,.13,.28}
\definecolor{burntorange}{rgb}{0.81,.33,0}

\ifdefined\ShowNotes
  \newcommand{\colornote}[3]{{\color{#1}\bf{#2: #3}\normalfont}}
\else
  \newcommand{\colornote}[3]{}
\fi

\newcommand {\note}[1]{\colornote{maroon}{Note}{#1}}
\newcommand {\todo}[1]{\colornote{cyan}{TODO}{#1}}
\newcommand {\lihi}[1]{\colornote{magenta}{LZ}{#1}}
\newcommand {\itamar}[1]{\colornote{blue}{IT}{#1}}
\newcommand {\roey}[1]{\colornote{red}{RM}{#1}}

\newcommand{\ignorethis } [1] {}
\newcommand{\DB         }     {{\mathcal{D}}}
\newcommand{\THR        }     {{\tau}}
\newcommand{\shortcite       }     {{\cite}}

\newenvironment{claim}[1]{\par\noindent{\bf Claim:}\space#1}{}
\newenvironment{claimproof}[1]{\par\noindent{\bf Proof:}\space#1}{\hfill $\blacksquare$}

%-----------------------------------------------------------------------------------
\begin{abstract}

We propose a novel measure for template matching named Deformable Diversity Similarity -- based on the diversity of feature matches between a target image window and the template. 
We rely on both local appearance and geometric information that jointly lead to a powerful approach for matching.
Our key contribution is a similarity measure, that is robust to complex deformations, significant background clutter, and occlusions. 
Empirical evaluation on the most up-to-date benchmark shows that our method outperforms the current state-of-the-art in its detection accuracy while improving computational complexity.
\end{abstract}

%-----------------------------------------------------------------------------------
 
\vspace*{-0.5cm}
\section{Introduction}
\label{sec:intro}

\begin{figure}
		\centering
		\rotatebox{90}{\hspace{0.1cm}}~
		\vspace{0.07cm}
		\begin{subfigure}[b]{.35\textwidth}\fcolorbox{CornflowerBlue}{yellow}{\includegraphics[width=.9\linewidth]{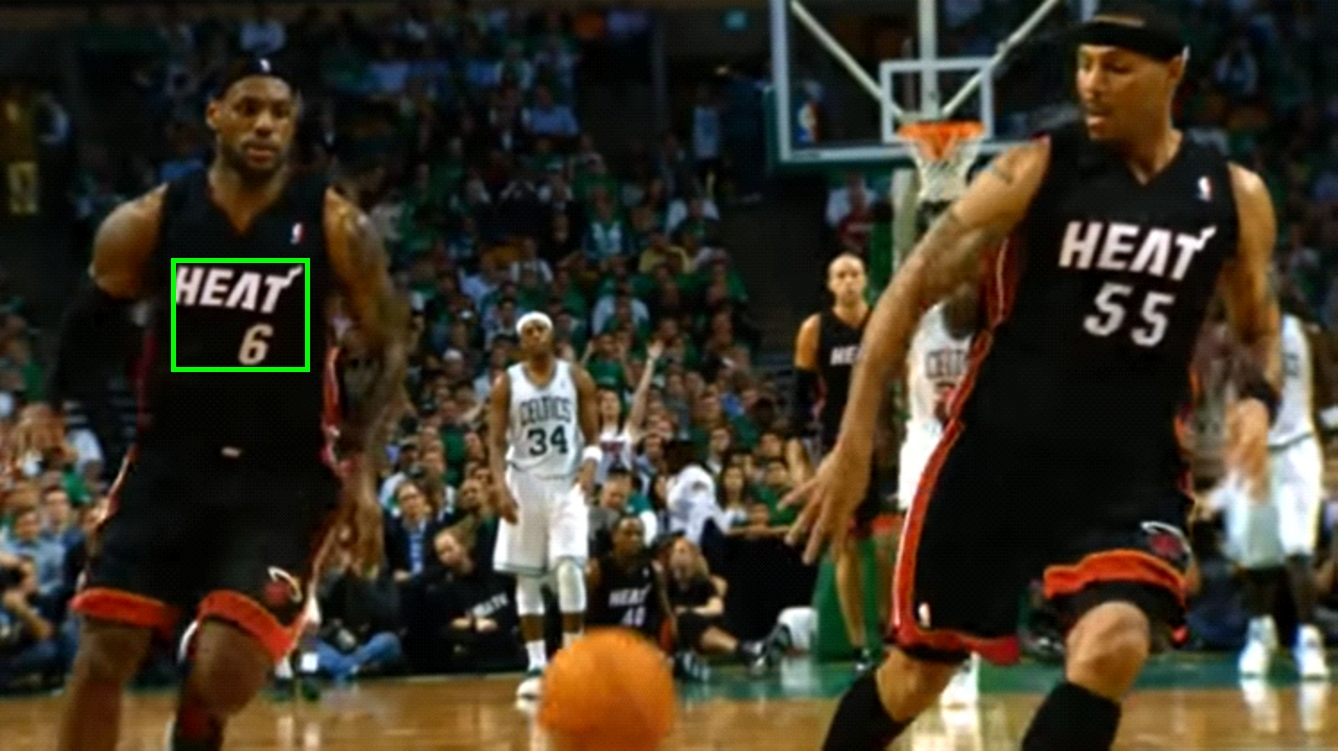}}\end{subfigure}
\centering
\rotatebox{90}{\hspace{0.1cm}}~
		\vspace{0cm}
		\begin{subfigure}[b]{.48\textwidth}\fcolorbox{green}{yellow}{\includegraphics[width=.9\linewidth]{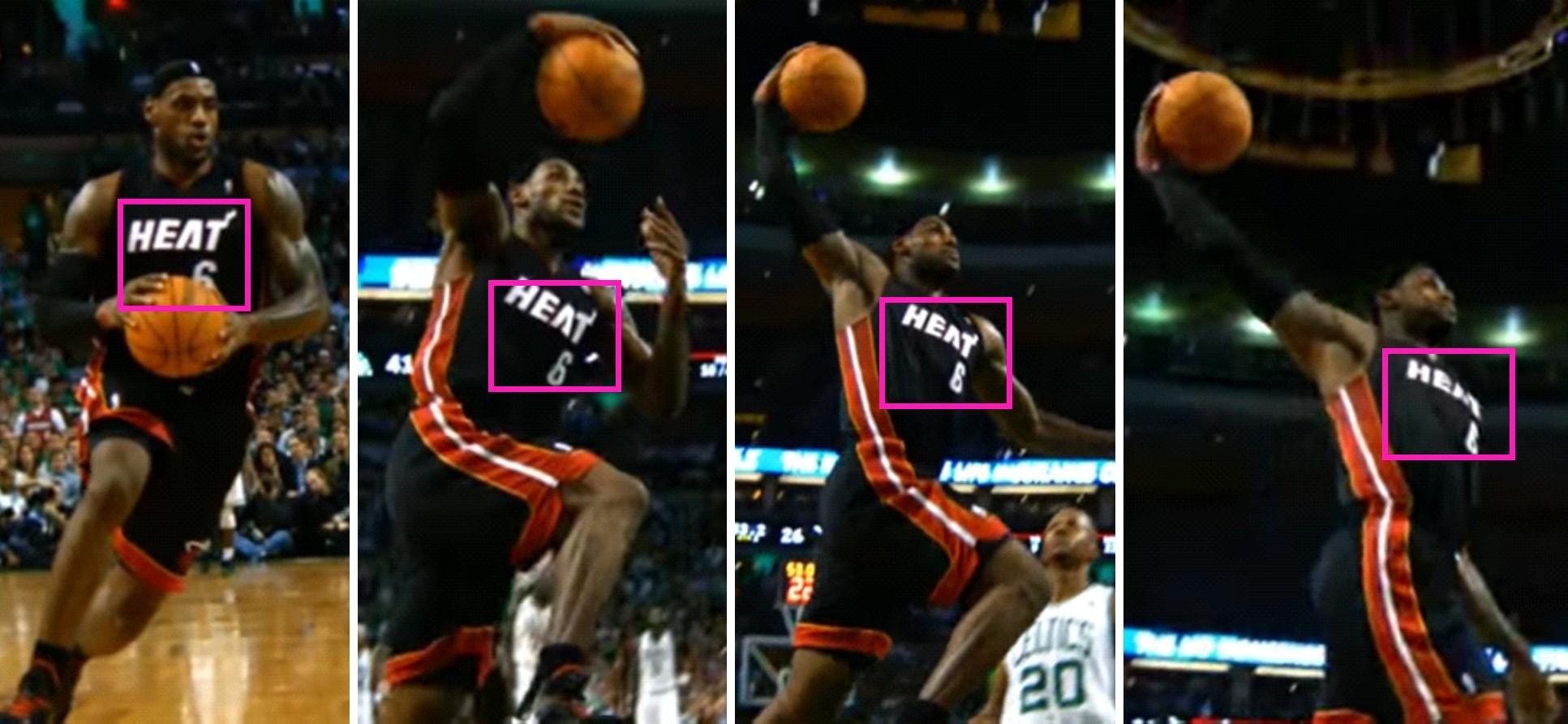}}\end{subfigure}~
% 		\begin{subfigure}[b]{.11\textwidth}\fcolorbox{green}{yellow}{\includegraphics[height=3.3cm]{Intro/ExampleNBA/I2c.jpg}}\end{subfigure}~
%         \begin{subfigure}[b]{.12\textwidth}\fcolorbox{CornflowerBlue}{yellow}{\includegraphics[height=3.3cm]{Intro/ExampleNBA/I3c.jpg}}\end{subfigure}~
%         \begin{subfigure}[b]{.14\textwidth}\fcolorbox{green}{yellow}{\includegraphics[height=3.3cm]{Intro/ExampleNBA/I4c.jpg}}\end{subfigure}
	\caption{ \textbf{Template Matching challenges:}  Template Matching results of the proposed Deformable Diversity Similarity (DDIS). (Top) The Miami Heats logo on Lebron James's shirt, is marked as template (in green). (Bottom) Best matches found by DDIS in four target images (in magenta). Different challenges are evident: the ball occludes the logo, Lebron moves non-rigidly with out-of-plane rotation and complex deformation of the shirt. 	
	}\vspace{-0.5cm}
	\label{fig:templateMatchingEx}
\end{figure}

Template Matching is a key component in many computer vision applications such as object detection, tracking, surveillance, medical imaging and image stitching. Our interest is in Template Matching ``in the wild''~\cite{dekel2015best}, i.e., when no prior information is available on the target image. An example application is to identify the same object in different cameras of a surveillance system~\cite{assari2016human}.
Another use case is in video tracking, where Template Matching is used to detect drifts and relocate the object after losing it~\cite{hong2015multi}. This is a challenging task when the transformation between the template and the target in the image is complex, non-rigid, or contains occlusions, as illustrated in Figure~\ref{fig:templateMatchingEx}. 

\fboxsep=0mm%padding thickness
\fboxrule=0pt%border thickness

Traditional template matching approaches, such as Sum-of-Squared-Distances or Normalized Cross-Correlation, do not handle well these complex cases. This is largely because they penalize all pixels of the template, which results in false detections when occlusions or large deformations occur. To overcome this limitation the Best-Buddies-Similarity (BBS) measure was proposed in~\cite{dekel2015best,oron2016best}. BBS is based on properties of the Nearest-Neighbor (NN) matches beween features of the target and features of the template. 
It relies only on a subset of the points in the template, thus hoping to latch on to the relevant features that correspond between the template and the target. This makes BBS more robust than previous methods.

In this paper we adopt the feature-based, parameter-free, approach of BBS and propose a novel similarity measure for template matching named DDIS: \emph{Deformable Diversity Similarity}. DDIS is based on two properties of the Nearest Neighbor field of matches between points of a target window and the template.
The first is that the \emph{diversity} of NN matches forms a strong cue for template matching. 
This idea is supported by observations in~\cite{jamrivska2015lazyfluids}, where patch diversity was used to match objects for texture synthesis. We propose formulas for measuring the NN field diversity and further provide theoretical analysis as well as empirical evaluations that show the strength of these measures.

The second key idea behind DDIS is to explicitly consider the \emph{deformation} implied by the NN field. As was shown by the seminal work of~\cite{felzenszwalb2010object} on Deformable Part Models, allowing deformations while accounting for them in the matching measure is highly advantageous for object detection. 
DDIS incorporates similar ideas for template matching leading to a significant improvement in template detection accuracy in comparison to the state-of-the-art.

A benefit of DDIS with respect to BBS~\cite{dekel2015best,oron2016best} is reduced computational complexity. Both measures rely on NN matches, however, BBS is formulated in a way that requires heavier computations. DDIS is more efficient while providing statistical properties similar to BBS.

To summarize, in this paper we introduce DDIS, a measure for \emph{template matching in the wild} that relies on two observations: (i) The diversity of NN matches between template points and target points is indicative of the similarity between them. (ii) The deformation implied by the NN field should be explicitly accounted for. DDIS is robust and parameter free, it operates in unconstrained environments and shows improved accuracy compared to previous methods on a real challenging data-set. Our code is available at \url{https://github.com/roimehrez/DDIS}

\section{Related Work}
\label{sec:related}

The similarity measure between the template and a sub-window of the target image is the core part of template matching. 
A good review is given in~\cite{ouyang2012performance}.
The commonly used methods are pixel-wise, e.g., Sum of Squared differences (SSD), Sum of Absolute Differences (SAD) and Normalized Cross-Correlation (NCC), all of which assume only translation between the template and target. 
They could be combined with tone mapping to handle illumination changes~\cite{hel2014matching} or with asymmetric correlation to handle noise~\cite{elboher2013asymmetric}. To increase robustness to noise pixel-wise measures such as M-estimators~\cite{chen2003fast,sibiryakov2011fast} or Hamming-based distances~\cite{shin2007fast,pele2008robust} have been proposed. 

More general geometric transformation such as affine are addressed by~\cite{tsai2002rotation,kim2007grayscale}. In~\cite{korman2013fast} parametric transformations are handled by approximating the global optimum of the parametric model. In~\cite{tian2012globally} non-rigid transformations are addressed via parametric estimation of the distortion. All of these methods work very well when their underlying geometric assumptions hold, however, they fail in the presence of complex deformations, occlusions and clutter.  

A second group of methods consider a global probabilistic property of the template. For example in~\cite{comaniciu2000real,perez2002color} color Histogram Matching is used (for tracking). This does not restrict the geometric transformation, however, in many cases the color histogram is not a good representation, e.g., in the presence of background clutter and occlusions. 
Other methods combine geometric cues with appearance cues. For example, a probabilistic solution was suggested in~\cite{olson2002maximum}, where geometric and color cues are used to represent the image in the location-intensity space. Oron et al.~\cite{oron2015locally} extend this idea by measuring one-to-one distance in $xyRGB$ space. These methods all make various assumptions that do not hold in complex scenarios.

A more robust approach, that can handle complex cases has been recently suggested in~\cite{dekel2015best,oron2016best}. Their approach, named the Best-Buddies-Similarity (BBS) is based on the Bi-Directional Similarity (BDS) concept of~\cite{simakov2008summarizing}. They compute the similarity between a template and a target window by considering matches between their patches. The matches are computed in both directions providing robustness to outliers. A similar idea was suggested in~\cite{dubuisson1994modified} by replacing the max operator of the Hausdorff distance~\cite{huttenlocher1993comparing} with a sum.
The BBS of~\cite{dekel2015best,oron2016best} lead to a significant improvement in template matching accuracy over prior methods. In this paper we propose a different measure, that shares with BBS its robustness properties, while yielding even better detection results.

\section{Diversity as a Similarity Measure} 
\label{sec:DIS}

\begin{figure}
	\captionsetup[sub]{skip=0.2pt}
		\centering
		\rotatebox{90}{\hspace{0.1cm}}~
		\vspace{0.1cm}
		\begin{subfigure}[b]{.23\textwidth}{\caption{}\includegraphics[width=.98\linewidth]{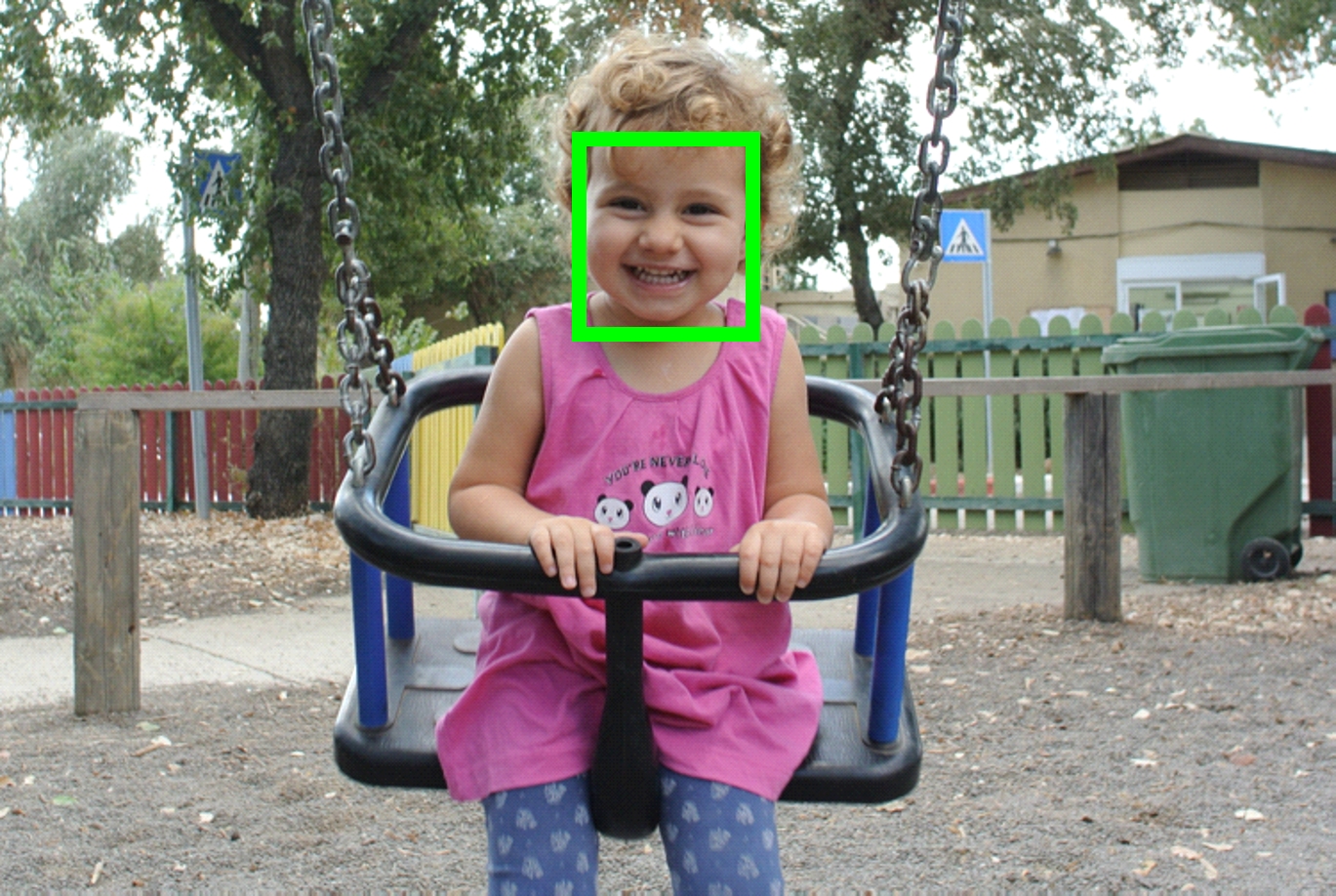}}\end{subfigure}~
		\begin{subfigure}[b]{.23\textwidth}{\caption{}\includegraphics[width=.98\linewidth]{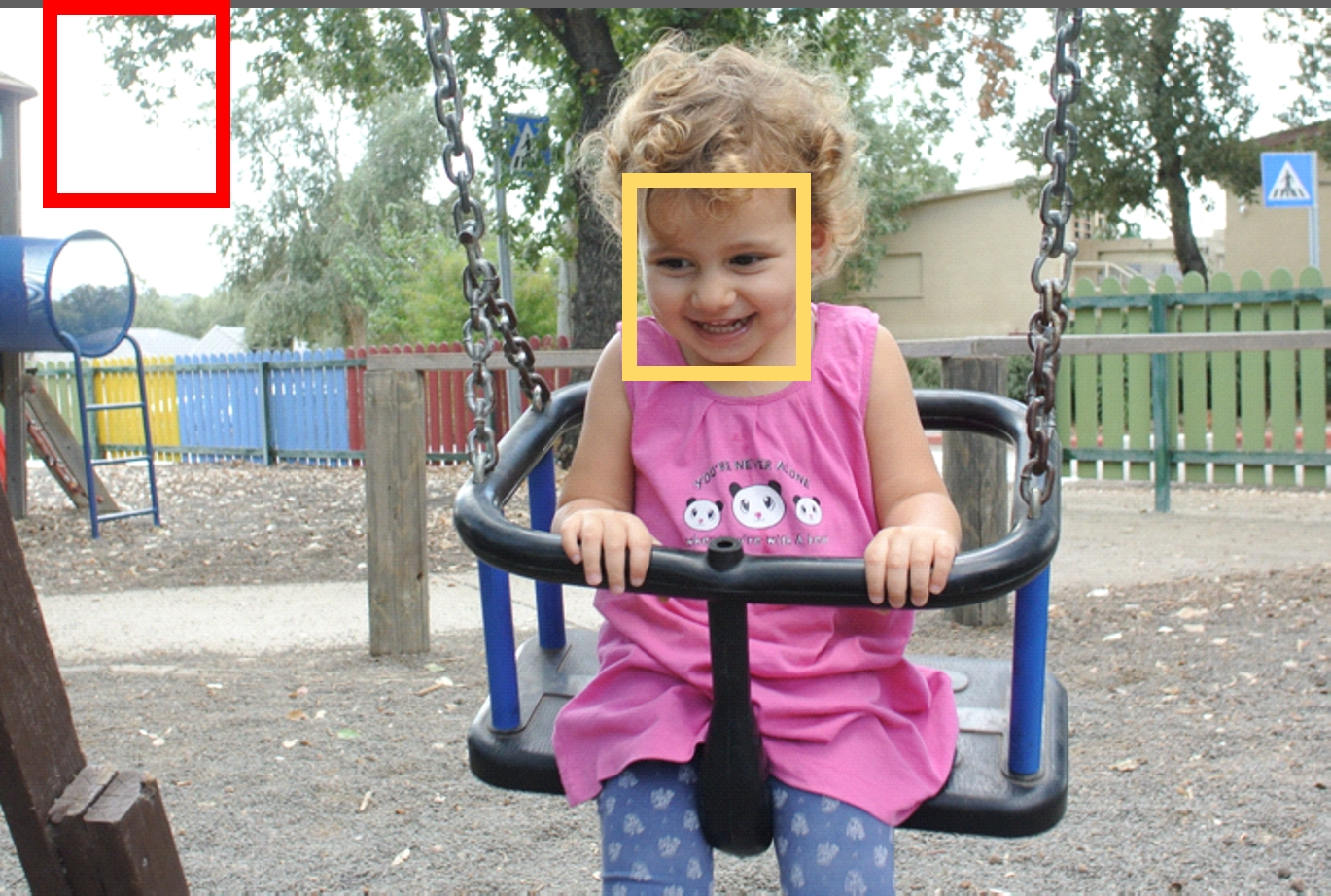}}\end{subfigure}
		\centering
        \rotatebox{90}{\hspace{0.1cm}}~
		\vspace{-0.1cm}
        \begin{subfigure}[b]{.15\textwidth}\fcolorbox{green}{green}{\includegraphics[width=.98\linewidth]{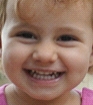}}\caption{}\label{fig:diversity:template}\end{subfigure}~
        \begin{subfigure}[b]{.15\textwidth}\fcolorbox{yellow}{yellow}{\includegraphics[width=.98\linewidth]{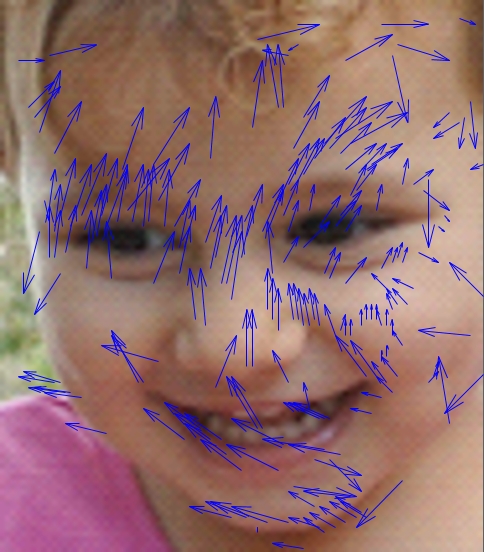}}\caption{}\label{fig:diversity:similar}\end{subfigure}~
        \begin{subfigure}[b]{.15\textwidth}\fcolorbox{Red}{Red}{\includegraphics[width=.98\linewidth]{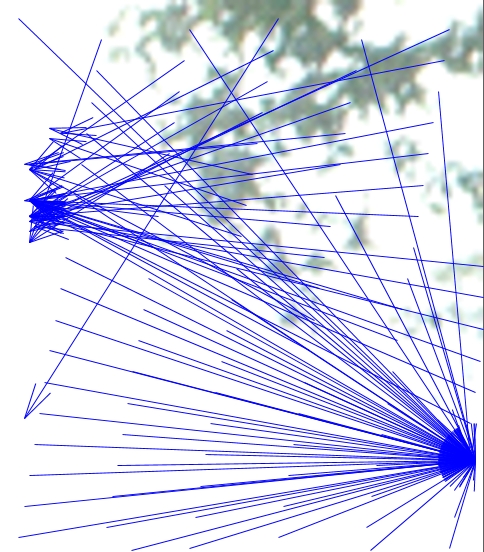}}\caption{}\label{fig:diversity:dissimilar}\end{subfigure}~
	\caption{{\bf Deformable Diversity:} A face template (c), marked in green (a), is searched in a target image (b). The face undergoes a complex transformation: background clutter, out of plain rotation, and non-rigid deformation. We zoom on a bad target window (e) and on the best target window (d), both are also marked by color in (b). 
The blue arrows are samples of the NN field: their start point marks a target patch while the end point marks the position of its NN in the template. The bad target (e), shows low diversity of NN patches with large deformation (long arrows). Conversely, the best target (d) shows high diversity and small deformation (short arrows). Best viewed on screen.}
    \vspace{-0.1cm}
	\label{fig:diversity}
\end{figure}

To measure similarity between a target window and a template we first find for every target patch its Nearest Neighbor (NN), in terms of appearance, in the template. 
Our key idea is that the similarity between the target and the template is captured by two properties of the implied NN field.
First, as shown in Figure~\ref{fig:diversity:similar}, when the target and template correspond, most target patches have a unique NN match in the template. This implies that the NN field is highly \emph{diverse}, pointing to many different patches in the template.  
Conversely, as shown in Figure~\ref{fig:diversity:dissimilar}, for arbitrary targets most patches do NOT have a good match, and the NNs converge to a small number of template points that happen to be somewhat similar to the target patches. 
Second, we note that arbitrary matches typically imply a large \emph{deformation}, indicated by long arrows in Figure~\ref{fig:diversity:dissimilar}. 

Next, we propose two ways for quantifying the amount of diversity and deformation of the NN field. The first is more intuitive and allows elegant statistical analysis. The second is slightly more sophisticated and more robust.

\fboxsep=0mm%padding thickness
\fboxrule=1pt%border thickness

%\vspace{-0.1cm}
%\paragraph{Diversity Similarity (DIS):}
\subsection{Diversity Similarity (DIS)}
Let points $p_i,q_j \in \mathbb{R}^{d}$ represent patches of the template and target, respectively. Our goal is to measure the similarity between two sets of points, the template points $P\!=\!\{p_i\}_{i=1}^{N}$ and the target points $Q\!=\!\{q_j\}_{j=1}^{M}$.  
We require finding the NN in $P$ for every point $q{\in}Q$, s.t., $\mathrm{NN}(q,P)=\argmin_{p{\in}P} d(q,p)$ for some given distance function $d(q,p)$.
The first property our measures are based on is the diversity of points $p{\in}P$ that were found as NNs. 

An intuitive way to measure diversity is to count the number of unique NNs. We define the \emph{Diversity Similarity} (DIS) as:
\begin{equation}
\label{eq:DIS}
\underset{Q{\rightarrow}P}{\textbf{DIS}} =
c \  \left|\left\{p_i\!\in\! P\! : \exists \: q_j\!\in\! Q\!,\: \mathrm{NN}(q_j,P)\!=p_i\!\right\} \right|
\end{equation}
where $c\!=\!{1}\!\slash \min{\{M,N\}}$ is a normalization factor and $\left|\{\cdot\}\right|$ denotes group size.

To provide further intuition as to why DIS captures the similarity between two sets of points we provide an illustration in 2D in Figure~\ref{fig:intuitiveExample}. 
Figure~\ref{fig:intuitiveExample:PQsame} demonstrates that when the distributions of points in $P$ and $Q$ are similar, most of the points $q{\in}Q$ have a unique NN $p{\in}P$ implying a high DIS value. 
Conversely, when $P$ and $Q$ are distributed differently, as illustrated in
Figure~\ref{fig:intuitiveExample:pQqP}, DIS is low. 
This is since in areas where $Q$ is sparse while $P$ is dense most of the points in $P$ are not NN of any $q$.
In addition, in areas where $Q$ is dense while $P$ is sparse most of the points in $Q$ share the same NNs.
In both cases, since the number of points is finite, the overall contribution to DIS is low. 

%As we show later in the experiments in Section~\ref{sec:experiments}, representing patches in $xyRGB$ and using DIS for template matching leads to results comparable to the state of the art.

%\paragraph{Deformable Diversity Similarity (DDIS):}
\subsection{Deformable Diversity Similarity (DDIS)}
While capturing well diversity, DIS does not explicitly consider the deformation field. 
Accounting for the amount of deformation is important since while non-rigid transformations should be allowed, they should also be restricted to give preference to plausible deformations of real objects. 

In order to integrate a penalty on large deformations we make two modifications to the way we measure diversity. First, to obtain an explicit representation of the deformation field we distinguish between the appearance and the position of each patch and treat them separately.
Second, we propose a different way to measure diversity, that enables considering the deformation amount.

Let $p^a$ denote the appearance and $p^l$ the location of patch $p$ (and similarly for $q$). 
We find the appearance based NN $p_i$ for every point $q_j$ s.t.  $p_i\!=\!\mathrm{NN}^a(q_j,P)=\argmin_{p\in P}d(q_j^a,p^a)$ for some given distance $d(q^a,p^a)$.
The location distance between a point $q_j$ and its $\mathrm{NN}^a$ is denoted by $r_j\!=\!d(q_j^l, p_i^l)$.
To quantify the amount of diversity as a function of the NN field we define $\kappa(p_i)$ as the number of patches $q{\in}Q$ whose $\mathrm{NN}^a$ is $p_i$:
\begin{equation}
\label{eq:kappa}
\kappa(p_i) = \left|\left\{q\in Q : \mathrm{NN}^a(q,P)=p_i\right\} \right|
\end{equation}
Finally, we define the \emph{Deformable Diversity Similarity} (DDIS) by aiming for high diversity and small deformation:
\begin{equation}
\label{eq:DDIS}
\underset{Q{\rightarrow}P}{\textbf{DDIS}}=c \ \sum_{q_j\in Q}\frac{1}{r_j\!+\!1} \cdot {\exp}\bigg({1-\kappa(\mathrm{NN}^a(q_j,P))}\bigg)
\end{equation}
where $c={1} \slash \min{\{M,N\}}$ is a normalization factor.

This definition can be viewed as a sum of contributions over the points $q_j$. When a point $q_j$ has a unique NN, then $\kappa(\mathrm{NN}^a(q_j,P))=1$ and the exponent reaches its maximum value of $1$. Conversely, when the NN of $q_j$ is shared by many other points, then $\kappa(\mathrm{NN}^a(q_j,P))$ is large, the exponent value is low and the overall contribution of $q_j$ to the similarity is low. In addition, the contribution of every point is inversely weighted by the length $r_j$ of its implied deformation vector.

DDIS possesses several properties that make it attractive: 
(1) it relies mostly on a subset of matches, i.e., points that have distinct NNs. Points that share NNs will have less influence on the score.
(2) DDIS does not require any prior knowledge on the data or its underlying deformation.
(3) DDIS analyses the NN field, rather than using the actual distance values.
These properties allow DDIS to overcome challenges such as background clutter, occlusions, and non-rigid deformations.

%\paragraph{DIS as simplified DDIS:}
\subsection{DIS as simplified DDIS}
DIS and DDIS capture diversity in two different ways. 
DIS simply counts unique matches in $P$, while DDIS measures exponentially the distinctiveness of each NN match of patches in $Q$. 
Nonetheless, we next show that DIS and DDIS are highly related. 

We start by ignoring the deformations by setting $r_j\!=\!0$ in~\eqref{eq:DDIS} and simplifying (without loss of generality) by assuming $M\!=\!N$.
We denote by $1/k$ the fraction of points $p{\in}P$ that are NNs of at least one point in $Q$.
Both DIS and DDIS reach their maximum value of $1$ when $k\!=\!1$, i.e., when $\kappa(p_i)\!=\!1 \ \forall p_i{\in}P$.
When $k\!=\!N$, i.e., all $q{\in}Q$ share a single NN, both scores reach their minimum value, DIS=$1/N$ and DDIS=$\exp(1-N)$. 
%\todo{DELETE: Between these extremum values both measures drop monotonically as a function of $k$. }

\begin{figure}[t]
	\captionsetup[subfigure]{labelformat=empty}
	\begin{center}
		\begin{subfigure}{.45\textwidth}
			\includegraphics[width=.98\linewidth]{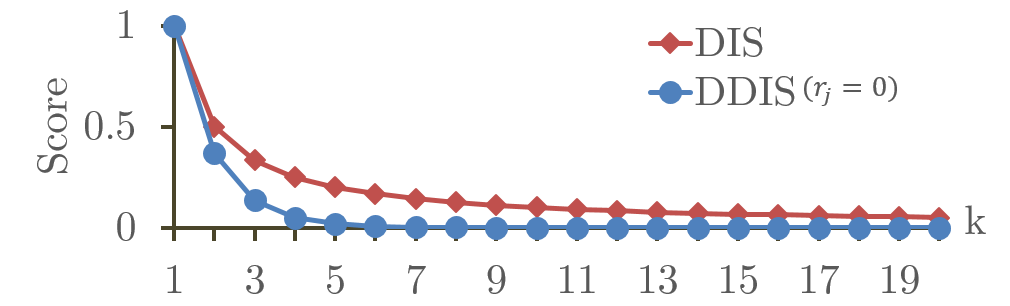}
		\end{subfigure}~
		\vspace{-0.5cm}
	\end{center}   
	\caption{\textbf{DIS as simplified DDIS:} Plots of DIS and DDIS (with $r_j\!=\!0$, $M\!=\!N$) as a function of $k$, where $N/k$ points in $P$ are NNs of some $q$, and for all of them $\kappa(p_i)\!=\!k$. Both DIS and DDIS are maximal at $k=1$ and decrease monotonically, reaching their minimal value when $k=N$. 
	}
	\vspace{-0.4cm}
	\label{fig:DDIS_vs_DIS}
\end{figure}

%\todo{DELETE; To see this we plot in Figure~\ref{fig:DDIS_vs_DIS} the values of DDIS and DIS as a function of $k$, assuming uniform distribution of NN matches.} 
Further intuition can be derived from the case of uniform distribution of NN matches, i.e., when only $N/k$ points in $P$ are NNs of some $q$, and for all of them $\kappa(p_i)\!=\!k$.
In this case $\mathrm{DIS}={(N/k)}/{N}\!=\!{1}/{k}$, and $\mathrm{DDIS}=\!{1}/{N} \sum_{q_j{\in}Q}{\exp}(1-k)\!=\!{\exp}(1-k)$. 
Both measures share extrema points between which they drop monotonically as a function of $k$, with DDIS decreasing faster due to its exponential nature. This is illustrated in Figure~\ref{fig:DDIS_vs_DIS}.
%\todo{SUGGESTING TO DELETE LAST SENTENCE: The exponential nature of DDIS with respect to the NN diversity is preferable over the linear nature of DIS. It makes DDIS a more robust measure of diversity than DIS.} 

%The difference in values is insignificant, since the score values are only used for comparing targets, hence, as long as monotonicity with respect to $k$ is maintained both scores will lead to the same results.

%=======================================================
\subsection{Statistical Analysis}
%=======================================================
To further cement our assertions that diversity captures the similarity between two distributions, we provide statistical analysis, similar to that presented in~\cite{dekel2015best,oron2016best}.
Our goal is to show that the expectation of DIS and DDIS is maximal when the points in both sets are drawn from the same distribution, and drops sharply as the distance between the two distributions increases.
We do that via a simple 1D mathematical model, in which an image window is modeled as a set of points drawn from a general distribution. 

Appendix~\ref{appendix:DIS_Expectation} presents derivations of $E\big[\mathrm{DIS}\big]$ (Expected value of DIS) when the points are drawn from two given distributions.
The expression for $E\big[\mathrm{DIS}\big]$ does not have a closed form solution, but it can be solved numerically for selected underlying distributions. Therefore, we adopt the same setup as~\cite{dekel2015best} where $P$ and $Q$ are assumed to be Gaussian distributions, which are often used as simple statistical models of image patches. We then use Monte-Carlo integration to approximate the Expectation for discrete choices of parameters $\mu_Q$ and $\sigma_Q$. For BBS and SSD we adopt the derivations in~\cite{dekel2015best}, where $E\big[\mathrm{BBS}\big]$ was also approximated via Monte-Carlo integration and $E\big[\mathrm{SSD}\big]$ is normalized.

\begin{figure}[t] 
	\captionsetup[sub]{font=scriptsize,justification=centering,skip=0.1pt}
    \centering
    \rotatebox{90}{\hspace{0.0cm}}~
    \vspace{0.05cm}
    \begin{subfigure}[b]{.15\textwidth}\caption{$E\big[\textbf{SSD}\big]$}\includegraphics[width=.98\linewidth]{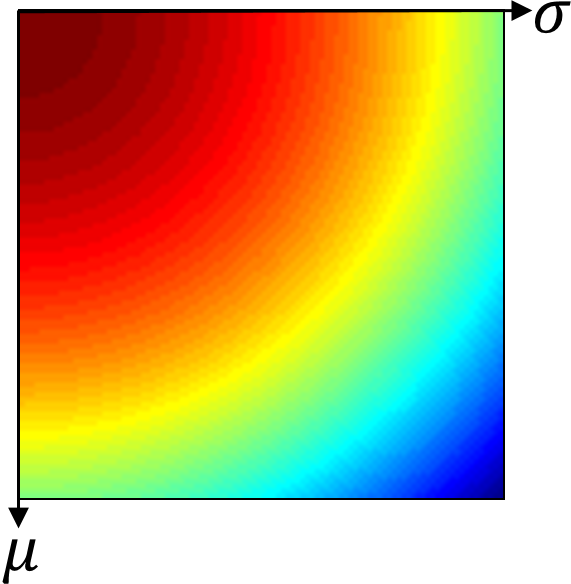}\end{subfigure}~
    \begin{subfigure}[b]{.15\textwidth}\caption{$E\big[\textbf{BBS}\big]$}\includegraphics[width=.98\linewidth]{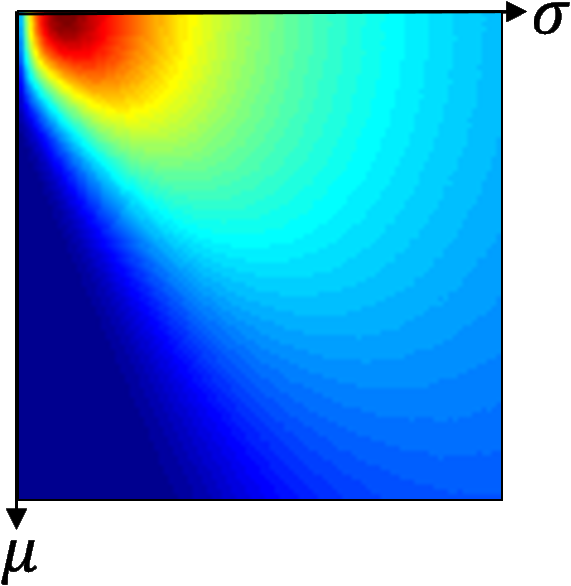}\end{subfigure}~
    \begin{subfigure}[b]{.15\textwidth}\caption{$E\big[\textbf{DIS}\big]$}\includegraphics[width=.98\linewidth]{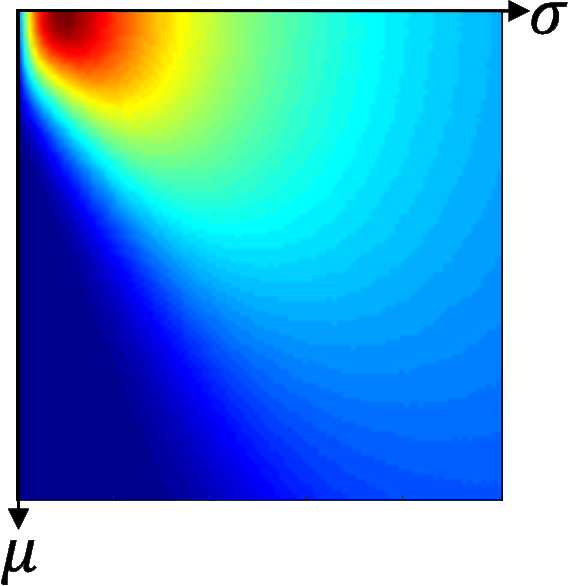}\end{subfigure}
    \rotatebox{90}{\hspace{0.0cm}}~
    \vspace{-0.05cm}
    \begin{subfigure}[b]{.15\textwidth}\includegraphics[width=.98\linewidth]{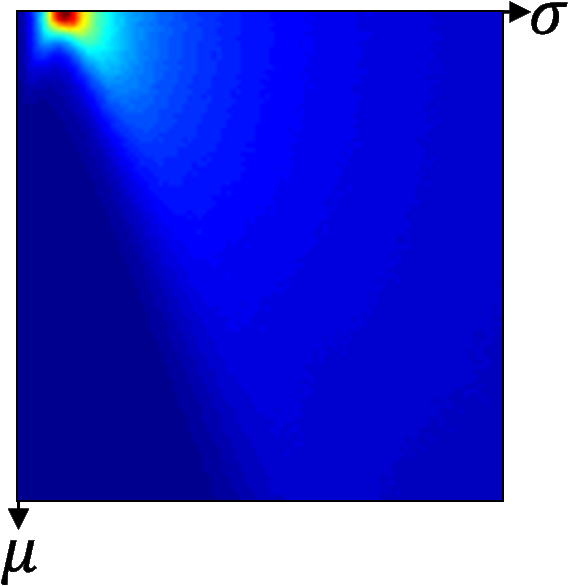}\caption{$E\big[\textbf{DDIS}\big]$\\\small{\emph{small} deformation}}\end{subfigure}~
    \begin{subfigure}[b]{.15\textwidth}\includegraphics[width=.98\linewidth]{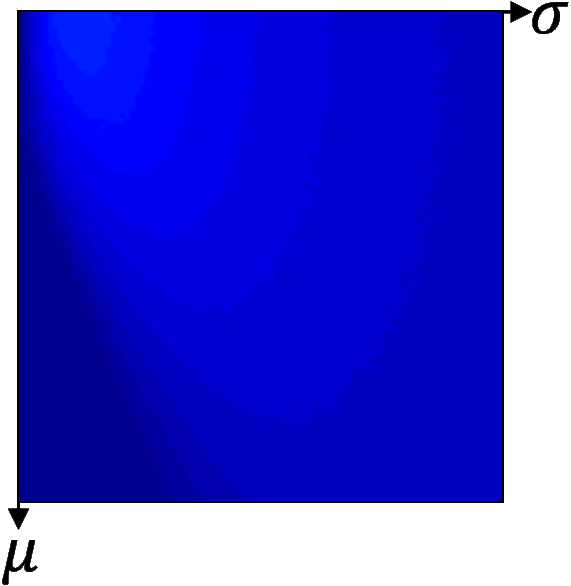}\caption{$E\big[\textbf{DDIS}\big]$ \\\small{\emph{large} deformation}}\end{subfigure}~
    \begin{subfigure}[b]{.15\textwidth}\includegraphics[width=.98\linewidth]{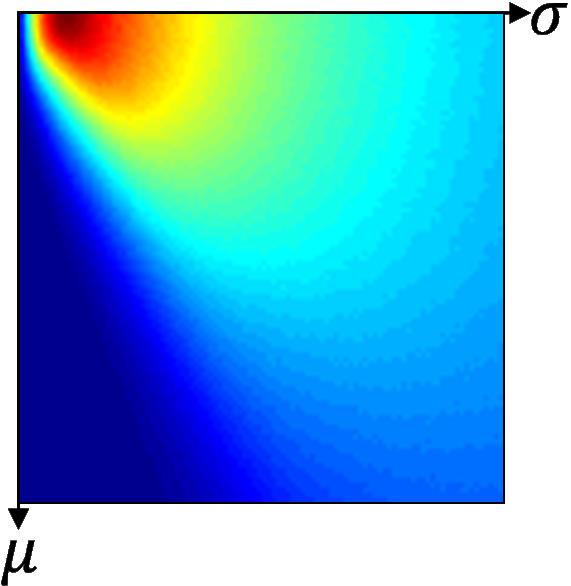}\caption{$E\big[\textbf{DDIS}\big]$ \\\small{\emph{ignore} deformation}}\end{subfigure}
%    \begin{subfigure}[b]{.165\textwidth}\includegraphics[width=.98\linewidth]{EDIS_minus_EBBS.png}\caption{$E\big[DIS\big]$-$E\big[BBS\big]$}\end{subfigure}~
	\caption{{\bf Expected behavior in the 1D Gaussian case:}
Following~\cite{dekel2015best} two point sets, $P$ and $Q$, are generated by sampling $N\!=\!M\!=\!100$ points from $N(0;1)$, and $N(\mu_Q;\sigma_Q)$, respectively, with $[\mu_Q,\sigma_Q]{\in}[0,10]$. 
(Top) The approximated expectation of SSD (a), BBS (b) and DIS (c) as a function of $\mu_Q$ and $\sigma_Q$ suggest that BBS and DIS behave similarly and drop much more rapidly than SSD as the distributions move apart. 
(Bottom) The approximated expectation of DDIS when the mean deformation is small (d), large (e), and ignored (f) ((d) and (e) are color scaled jointly).
It can be seen that small deformation fields correspond to a sharper peak in the expectation, while for large deformations the similarity is always low.
} \vspace{-0.4cm}
	\label{fig:ExpectationAnalysis}
\end{figure}

Figure~\ref{fig:ExpectationAnalysis} presents the resulting approximated expected values. It can be seen that DIS is likely to be maximized when the distributions are the same, and falls rapidly when the distributions differ from each other. In addition it is evident that DIS and BBS present highly similar behaviors. 
Finally, similar to~\cite{dekel2015best}, one can show that this holds also for the multi-dimensional case. 

For DDIS we cannot derive nice expressions for its Expectation $E\big[\mathrm{DDIS}\big]$. Instead, we use simulations to approximate it. 
The simulation needs to consider also locations to quantify the amount of deformation, $r_j$, in~\eqref{eq:DDIS}.
When $r_j=0$ the expectation is similar to that of BBS and DIS. 
For $r_j\neq0$ we simulate two cases: 
(i) Small deformation: We sort the points in each set based on their appearance coordinate, and take as position their index in the sorted list. 
When the distributions are different the diversity is very low anyhow. 
But when the distributions are similar, the sorting results in points and their NN having a \textit{similar index}, which corresponds to \textit{small deformation}. 
(ii) Large deformation: We sort the points of one set in descending order and the other set in ascending order, again taking as position their index in the sorted list. When the distributions are similar, the sorting results in points and their NN having a \textit{different index}, which corresponds to \textit{large deformation}. 
Figure~\ref{fig:ExpectationAnalysis} shows that for small deformation $E\big[\mathrm{DDIS}\big]$ drops sharply as the distributions become more different. For large deformations it is always low, as desired, since even when the appearances are similar, if the geometric deformation is large the overall similarity between the point sets is low. 

\section{Comparison to BBS} 
\label{sec:CompareToBBS}

Our measures bare resemblance to BBS -- all rely on NN matches between two sets of points. 
There are, however, two key differences: (i) the way in which similarity between the two sets is measured, and, (ii) the penalty on the amount of spatial deformation. 
We next analyze the implications of these differences.

\subsection{You Only Need One Direction}
\label{sec:youOnlyNeedOneDirection}

%\paragraph{Unilateral or Bilateral?}
The key idea behind the bi-directional similarity approaches of~\cite{dekel2015best,oron2016best,simakov2008summarizing} is that robust matching requires bi-directional feature correspondences. 
Our unilateral measures contradict this claim.
In fact, an observation we make is that, Diversity provides a good approximation to BBS.
The analysis we present is for DIS, since it is simpler than DDIS and does not incorporate deformation, making the comparison to BBS more fair and direct.

Recall that BBS counts the number of bi-directional NN matches between the target and template. A pair of points $p_i{\in}P$ and $q_j{\in}Q$ are considered a best-buddies-pair (BBP) if $p_i$ is the NN of $q_j$, and $q_j$ is the NN of $p_i$. BBS counts the number of BBPs as a measure of similarity between $P$ and $Q$.
Clearly, the bilateral nature of BBS is wasteful in terms of computations, compared to the unilateral DDIS and DIS.

%\begin{equation}
%\label{eq:BBS}
%\underset{P,Q}{\textbf{BBS}}=c \cdot \left|\left\{ (p_i{\in}P,q_j{\in}Q): {bb_{i,j}(P,Q)} \right\} \right|
%\end{equation}
%%
%%
%where $c={1} \slash \min{\{M,N\}}$  and $bb_{i,j}(P,Q)$ is the best buddies pair indicator:
%\begin{equation}
%\label{eq:bbij}
%bb_{i,j}(P,Q) =
%\begin{dcases*}
%\begin{array}{ll}
%1\! &\! \mathrm{NN}(p_i,Q)\!=q_j\! \wedge\! \mathrm{NN}(q_j,P)\!=p_i\!\\
%0\! &\!\text{otherwise}
%\end{array}
%\end{dcases*}
%\end{equation}

\begin{figure}[t]
    	\centering
		\rotatebox{90}{\hspace{0.1cm}}~
		\vspace{-0.2cm}
		\begin{subfigure}[b]{.19\textwidth}\includegraphics[height=2.15cm]{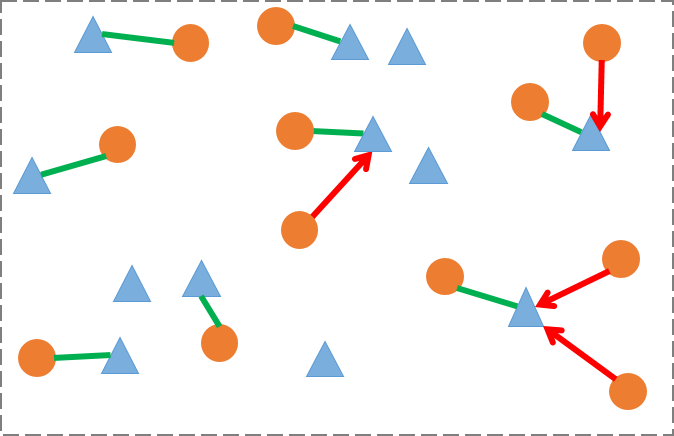}\caption{similar distributions}\label{fig:intuitiveExample:PQsame}\end{subfigure}~
        \begin{subfigure}[b]{.3\textwidth}\includegraphics[height=2.15cm]{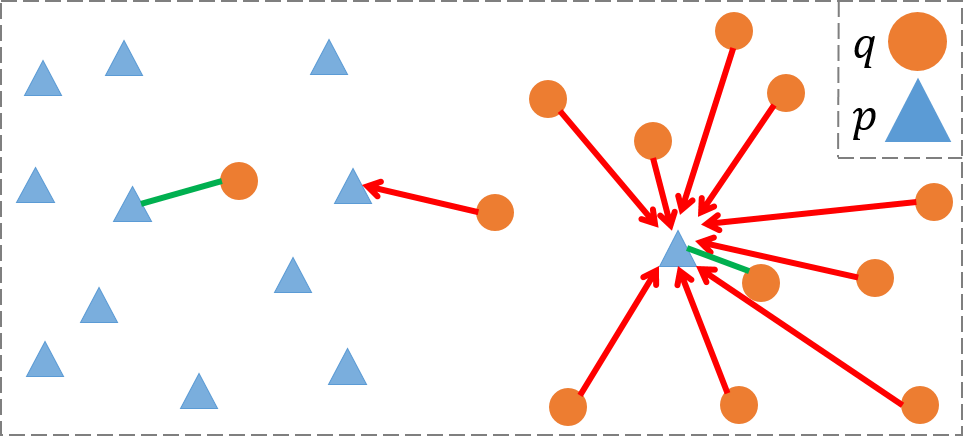}\caption{different distributions}\label{fig:intuitiveExample:pQqP}\end{subfigure}
	\caption{{\bf Intuition for DIS and BBS:}  
    For each point $q{\in}Q$ we draw a red arrow pointing at its NN $p \in P$. If $q$ and $p$ are also best-buddies-pair (BBP, see section ~\ref{sec:youOnlyNeedOneDirection}), we change the red arrow to a green line. DIS counts blue triangles that are pointed to by either a red arrow or a green line. BBS counts green lines. 
	(a) $P$ and $Q$ are similarly distributed, hence, many $p$'s are NN of some $q$ and there are many BBPs. Here $DIS\!=\!BBS\!=\!8$. 
	(b) $P$ and $Q$ have different distributions. A single $q$ among dense $p$'s or a single $p$ among dense $q$'s contribute $1$ to both DIS and BBS. Occasionally, there is a unique NN match between $Q$ and $P$ that is not a BBP. Since the distributions of $P$ and $Q$ are different both DIS and BBS are relatively low, $DIS\!=\!3$ and $BBS\!=\!2$.}
	\vspace{-0.5cm}
	\label{fig:intuitiveExample}
\end{figure}

DIS and BBS are defined differently, however, since the number of patches in both target and template is \emph{finite}, DIS provides a good approximation to BBS. 
As illustrated in Figure~\ref{fig:intuitiveExample:PQsame} when the distributions of points in $P$ and $Q$ are similar, many of the NN relations are bi-directional. This implies that the values of BBS and DIS are very similar. In the extreme case when the template and target are identical, every point $q$ has a unique NN $p\!=\!q$ and they form a BBP. In this case DIS=BBS exactly.

DIS and BBS behave similarly also when the distributions are different, as illustrated in Figure~\ref{fig:intuitiveExample:pQqP}. 
In areas where $P$ is sparse and $Q$ is dense we get multiple points $q{\in}Q$ that share the same NN $p{\in}P$. At most one of them forms a BBP and their joint contribution to both DIS and BBS is $1$.
Since the number of points in $P$ and $Q$ is finite, this implies that there are other areas where $P$ is dense and $Q$ is sparse. In these areas there are many points in $P$ that are not NN of any $q{\in}Q$, and have zero contribution to both DIS and BBS.

Our observations are in line with the Expectation analysis of Figure~\ref{fig:ExpectationAnalysis}. In addition, our experiments (Section~\ref{sec:experiments}) show that DIS and BBS achieve comparable template matching accuracy.

%A single $q$ in an area of many $p$'s has a single NN and by construction, they are necessarily also a BBP since $q$ is the NN of all the $p$'s in its vicinity. Therefore, the contribution to both DIS and BBS is $1$.
%%
%In addition, a single $p$ in an area dense with points $q$ is the NN of all the $q$'s, and only one of them is its BBP.
%Therefore, again the contribution to both DIS and BBS in only $1$ and both measures yield identical low scores.
%
%These observations can be generalized.

%In fact, we next prove that DIS bounds BBS from above.
%\input{YouOnlyNeedOneDirection/DIS_bounds_BBS.tex}
%
\subsection{Explicit vs. Implicit Deformation Modeling}
%\label{sec:explicit-vs-implicit}

%\paragraph{Explicit vs. implicit deformation modeling:}
%
The need to penalize large deformations was noted in~\cite{dekel2015best,oron2016best}. This was done implicitly by adding the $xy$ coordinates to the feature vectors when searching for NNs. The distance between a pair of points is taken as a weighted linear combination of their appearance and position difference.
This is different from DDIS that considers only appearance for NN matching and explicitly penalizes the deformation in the obtained NN field. 
Our approach has two benefits: (i) improved runtime, and (ii) higher detection accuracy. 

Using only appearance for NN matching significantly reduces runtime since while every image patch is shared by many sub-windows, its $xy$ coordinates are different in each of them. This implies that the NN field needs to be computed for each image sub-window separately. Conversely, working in appearance space allows us to perform a single NN search per image patch. 
In Section~\ref{sec:complexity} we analyze the benefits in terms of computational complexity.

Separating between appearance and position also leads to more accurate template localization. Overlapping target windows with very similar appearance could lead to very similar similarity scores. DDIS chooses the window implying less deformations. Our experiments indicated that this is important and improves the localization accuracy.

%=================================
%=================================
%============================
\section{Implementation}
\label{sec:Implementation}

To utilize DDIS for template matching in images, we follow the traditional raster scan approach. Our algorithm gets as input a target image $S$ and a template $T$. Its output is a frame placing $T$ within $S$. We denote the width of $T$ by $T_w$ and its height $T_h$, similarly for $S$. Each template sized sub-window $W{\in}S$ is compared to $T$. We extract from $T$ and $W$ feature vectors, as described below, yielding sets $P$ and $Q$ respectively. 
We use the Euclidean distance ($L2$) to compare appearance features  $p^a$ and $q^a$.
The deformation length $r_j$ is the Euclidean distance between the $xy$ coordinates $p^l$ and $q^l$.  
%, as illustrated in Figure~\ref{fig:nnf}
Our implementation consists of $4$ phases:

\textbf{0.} \emph{Feature extraction:}
We experimented with two forms of appearance feature, color and deep-features.
As color features we set $p^a$ and $q^a$ as vectorized RGB pixel values of $3{\times}3$ overlapping patches.
To obtain deep-features we used the popular VGG-Deep-Net~\cite{simonyan2014very}. More specifically, we take feature maps from layers ${conv1\_ 2}$, ${conv3\_ 4}$ and ${conv4\_ 4}$ (akin to the suggestion in~\cite{ma2015hierarchical} for object tracking). We forsake the higher layers since we found the low spatial resolution inaccurate. The features maps were normalized to zero mean and unit standard deviation, and then upscaled, via bi-linear interpolation, to reach the original image size.  

\textbf{1.} \emph{NN search:} We find for each feature vector in $S$, its approximate NN in the template $T$. We use TreeCANN~\cite{olonetsky2012treecann} with PCA dimensionality reduction to 9 dimensions, kd-tree approximation parameter $\epsilon=2$, dense search ($g_S=g_T=1$), and window parameter $w_S=3$, $w_T=5$. 

\textbf{2.} \emph{Similarity map calculation:} 
For each target image pixel (ignoring boundary pixels) we compute the similarity between its surrounding sub-window $W$ and the template $T$. For each $W$, we first compute $\kappa(p_i), \forall{p_i{\in}P}$ as defined in~\eqref{eq:kappa}. Since subsequent windows have many overlaps, the computation of $\kappa$ needs only update the removed and added features with respect to the previous sub-window. We then calculate DDIS as defined in~\eqref{eq:DDIS}.

\textbf{3.} \emph{Target localization:} 
Finally, the template location is that with maximum score. 
%Similarly to~\cite{dekel2015best}, when using deep feature, we take the maximum over the similarity map. 
Before taking the maximum, we smooth the similarity map with a uniform kernel of size $\frac{T_w}{3}{\times}\frac{T_h}{3}$, to remove spurious isolated peaks.

%========================================
%========================================
%========================================
\section{Complexity}
%========================================
\label{sec:complexity}
The sets $P$ and $Q$ consist of features from all locations in $T$ and $W$, receptively. This implies $|P|=|Q|=T_w \cdot T_h \triangleq l$. The number of possible sub-windows\footnote{In practice, we exclude patches that are not fully inside the template or the sub-window, but these are negligible for our complexity analysis.} is $S_w{\cdot}S_h \triangleq L$.
Recall that $d$ denotes the feature vectors length. For color features $d$ equals the size of the patch ${\times}3$ while for deep-features it is determined by the feature map dimension. 
Next, we analyze the complexity of steps (1-3).

\textbf{1.} \emph{NN search:} 
TreeCANN consists of two stages.
In the first, the dimension of all template points is reduced from $d$ to $d'$ in $O(dl)$ and a k-d tree is built in $O(d'l{\log}l)$. 
The second stage performs the queries. 
Each query consists of dimensionality reduction $O(d)$, a search in the k-d tree $O({\log}l)$ (on average), and a propagation phase which leverages spatial coherency $O(d)$. The overall complexity for finding the Approximate NN for all the features in the target image $S$ is $O\big(d'l{\log}l+L(d+{\log}l)\big)$ on average. The memory consumption is $O(l)$. 

\textbf{2.}
\emph{Similarity map calculation:} 
Assuming for simplicity that $T_w=T_h=\sqrt{l}$, the update of $\kappa(p)$ takes $O(\sqrt{l})$ operations, for any $W$ except for the first one. Next, DDIS is calculated with $O(l)$ operations. Given that the overall number of sub-windows is $L$ this step's complexity is $O\Big(L(l+\sqrt{l})\Big){\equiv}O\Big(Ll\Big)$. The memory consumption for this stage is $O(l)$ which is the size of a table holding $\kappa(p)$.

\textbf{3.}
\emph{Target localization:} 
Averaging the similarity map is done efficiently using an integral image in $O(L)$. To find the maxima location another swipe over the image is needed, which takes $O(L)$. 

Putting it all together, we get that the overall complexity of Template Matching with DDIS is 
%\begin{equation}
$O(d'l{\log}l+Ll)$
%\end{equation}
%
where we omitted $O(L(d+{\log}l))$ since $d$ and $l$ are expected to be of the same order for small $T$ and $d{\ll}l$ for large $T$.

%==============================
\vspace{-0.3cm}
\paragraph{Comparison to BBS}
One of the benefits of DDIS with respect to BBS is that it requires only unilateral matches. The benefit in terms of complexity can now be made clear. According to~\cite{oron2016best}, the BBS complexity using deep features is $O(Ll^4d)$ and for color features it is $O(Ll^2d)$ on average. The latter case uses heavy caching which consumes $O(l^2\sqrt{L})$ memory (assuming $S_w=S_h=\sqrt{L}$).

%In such comparison we want to show how the time changes when the size of $S$ and $T$ changes. For the motorcycle example, figure~\ref{fig:TM_examples}, $T$ is of size $54\times53$ and $S$ is of size $270\times480$ and the runtimes are $0.75$ and $8.7$ seconds for DDIS and BBS respectively.
%For The Cavs logo example, $T$ is of size $98\times100$ while $S$ is of size $324\times576$. DDIS runtime grows by $2.8$ on that example to $2.1$ seconds while BBS runtime grows by $24$ to $3.5$ minutes.

%==============================================================================================
\section{Empirical Evaluation} 
\label{sec:experiments}

Our experimental setup follows that of~\cite{oron2016best} that created a benchmark by sampling frames from  video sequences annotated with bounding-boxes for object tracking~\cite{wu2013online}. The videos present a variety of challenges: complex deformations, luminance changes, scale differences, out-of-plane rotations, occlusion and more. The benchmark consists of three data-sets, generated by sampling $\left\{270, 270, 254\right\}$ pairs of frames with a constant frame (time) difference $dFrame\!=\!\left\{25, 50, 100\right\}$, producing increasingly challenging data-sets, and overall, a challenging benchmark for template matching. 

For each pair of frames, one is used to define the template as the annotated ground-truth box, while the second is used as a target image. As commonly done in object tracking, the overlap between the detection result and the ground-truth annotation of the target is taken as a measure of accuracy: $Accuracy=\frac{|R_{est} \cap R_{truth}|}{|R_{est} \cup R_{truth}|}$ where $|\cdot|$ counts the number of pixels in a region and $R_{truth}$ and $R_{est}$ are the ground truth and estimated rectangles, locating $T$ in $S$. 

\definecolor{a}{rgb}{0.368627450980392,	0.309803921568627,	0.635294117647059}
\definecolor{b}{rgb}{0.218964460784314,	0.449052742700341,	0.725769761029412}
\definecolor{c}{rgb}{0.217039377734565,	0.593806353743544,	0.726823529411765}
\definecolor{d}{rgb}{0.368979906011992,	0.740264734090567,	0.651823529411765}
\definecolor{e}{rgb}{0.533184514766650,	0.819331026528258,	0.645061840120664}
\definecolor{f}{rgb}{0.703315298253005,	0.880247981545559,	0.640422322775264}
\definecolor{g}{rgb}{0.866313642850232,	0.950749711649366,	0.603092006033183}
\definecolor{h}{rgb}{0.924680708657847,	0.959019525779062,	0.632620251225490}
\definecolor{i}{rgb}{0.950000000000000,	0.950000000000000,	0.711568627450980}
\definecolor{j}{rgb}{0.985633161956189,	0.914789408852483,	0.622025774202887}
\definecolor{k}{rgb}{0.995796568627451,	0.843437860093870,	0.502511384216209}
\definecolor{l}{rgb}{0.993000919117647,	0.710557488742431,	0.398341248410999}
\definecolor{m}{rgb}{0.980465686274510,	0.553937172695165,	0.304411764705882}
\definecolor{n}{rgb}{0.948529474107957,	0.401903516759713,	0.264767156862745}
\definecolor{o}{rgb}{0.873591327519380,	0.291187047735761,	0.302450980392157}
\definecolor{p}{rgb}{0.767364911080711,	0.160386454929194,	0.302462469362745}
\definecolor{q}{rgb}{0.619607843137255,	0.0039215686274509,	0.258823529411765}

\begin{figure*}[h]
		\centering
		\rotatebox{90}{\hspace{0.001cm}}~
		\vspace{-0.2cm}
		\begin{subfigure}[b]{.25\textwidth}\includegraphics[width=\linewidth]{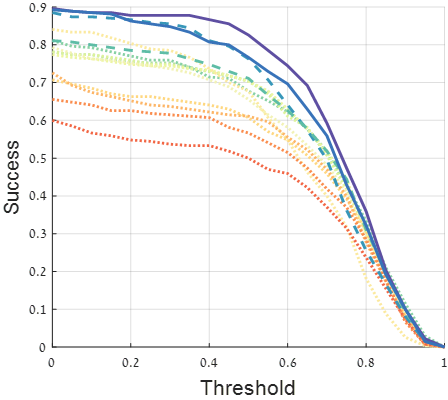}\caption{$dFrame=25$}\end{subfigure}~
        \begin{subfigure}[b]{.25\textwidth}\includegraphics[width=\linewidth]{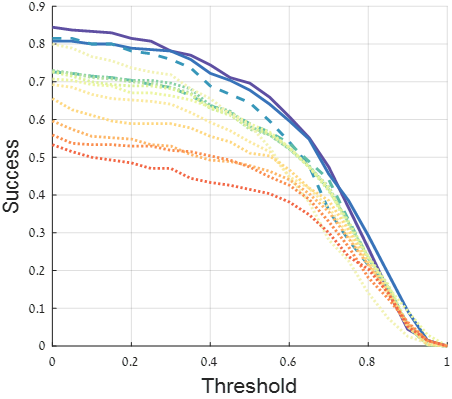}\caption{$dFrame=50$}\end{subfigure}~
        \begin{subfigure}[b]{.25\textwidth}\includegraphics[width=\linewidth]{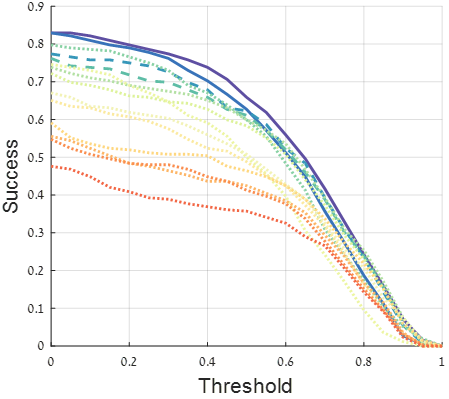}\caption{$dFrame=100$}\end{subfigure}~
        \begin{subfigure}[b]{.24\textwidth}
\begin{center}
\scalebox{0.62}{
\tabcolsep=0.13cm
\begin{tabular}{ l | c c c | r}
\multirow{2}{*}{\textbf{Method}} & \multicolumn{3}{c|}{\textbf{Dataset}} & \multirow{2}{*}{\textbf{Mean}}\\
%\cdashline{2-4}
	& $\textit{25}$ 		& $\textit{50}$ 	& $\textit{100}$ 	&	\\
\hline
\cellcolor{a!65}{DDIS - D}   & \textbf{0.679} & \textbf{0.583} & \textbf{0.571} &\textbf{0.611}\rule{0pt}{2ex}\\
\cellcolor{b!65}{DDIS - C}   & 0.650 & 0.590 & 0.540 &0.593\\
\hdashline
\cellcolor{c!65}{DIS - D}   & 0.630 & 0.549 & 0.518 &0.566\rule{0pt}{2ex}\\
\cellcolor{d!65}{BBS - D}   & 0.598 & 0.514 & 0.532 &0.548\\
\cellcolor{e!65}{SSD - D}   & 0.584 & 0.512 & 0.519 &0.538\\
\cellcolor{f!65}{SAD - D}   & 0.582 & 0.507 & 0.513 &0.534\\
\cellcolor{g!65}{NCC - D}   & 0.581 & 0.509 & 0.491 &0.527\\
\cellcolor{h!65}{BBS - C}   & 0.590 & 0.505 & 0.445 &0.513\\
\cellcolor{i!65}{DIS - C}   & 0.561 & 0.501 & 0.446 &0.503\\
\cellcolor{j!65}{BDS - C}   & 0.564 & 0.471 & 0.425 &0.486\\
\cellcolor{k!65}{BDS - D}   & 0.513 & 0.447 & 0.401 &0.454\\
\cellcolor{l!65}{SAD - C}   & 0.516 & 0.412 & 0.365 &0.431\\
\cellcolor{m!65}{NCC - C}   & 0.483 & 0.398 & 0.359 &0.413\\
\cellcolor{n!65}{SSD - C}   & 0.427 & 0.363 & 0.308 &0.366\\ 
\hline
\multicolumn{5}{c}{\large{\rule{0pt}{2.5ex}\textbf{D}=deep features}}\\
\multicolumn{5}{c}{\large{\textbf{C}=color space}}\\
\end{tabular}
}
%\vspace{0.5cm}
\caption{AUC results}        
\label{fig:auc:table}
\end{center}
\end{subfigure}~
    \caption{\textbf{Template matching accuracy}: Evaluation on the benchmark of \cite{oron2016best}: 270 template-image pairs with $dFrame\in\left\{25,50,100\right\}$. 
DDIS outperforms competing methods as can be seen in the ROC curves (a)-(c) showing the fraction of image-pairs with $Accuracy\!>\!Threshold{\in}[0,1]$. The corresponding Area-under-curve (AUC) scores are presented in the table in (d).
For all methods $D$ stands for deep features while $C$ stands for color. DDIS with deep features provides the best results. DDIS with color features comes in second, outperforming other methods even when they use deep-features.}   
\vspace{-0.5cm} 
\label{fig:auc}
\end{figure*}

\textbf{Quantitative Evaluation:} We compare DDIS and DIS to BBS, BDS, SSD, SAD and NCC with both color and deep features. For BBS and DIS we use the exact same setup as in~\cite{oron2016best}, that is, $3{\times}3$ non-overlapping patches represented in $xyHSV$ space. 
In Figure~\ref{fig:auc} we plot for each data-set and method a success rate curve.
It can be seen from the Area-Under-Curve (AUC) scores in Table~\ref{fig:auc:table} that DDIS is significantly more successful than all previous methods.
Furthermore, DDIS with our simplistic color features outperforms all other methods with either color or deep features.
When using Deep features, DDIS improves over BBS with margins of $\approx 11\%, 15\%, 8\%$ for the three data-sets. When using color features the margins are $\approx 10\%, 16\%, 21\%$.

\textbf{Qualitative Evaluation:} Figure~\ref{fig:TM_examples} displays several detection results on challenging examples, taken from the web, that include occlusions, significant deformations, background clutter and blur. We compare DDIS and DIS to BBS -- the current state-of-the-art. It is evident from the detection likelihood maps that DIS and BBS share a similar behavior, supporting our suggestion that unilateral matching suffices to capture similarity. DDIS, on the other hand, accounts also for deformations, hence, it presents cleaner maps, with fewer distractors.

\begin{figure}[b]
	\captionsetup[subfigure]{labelformat=empty}
	\begin{center}
		\begin{subfigure}{.48\textwidth}
			\includegraphics[width=.98\linewidth]{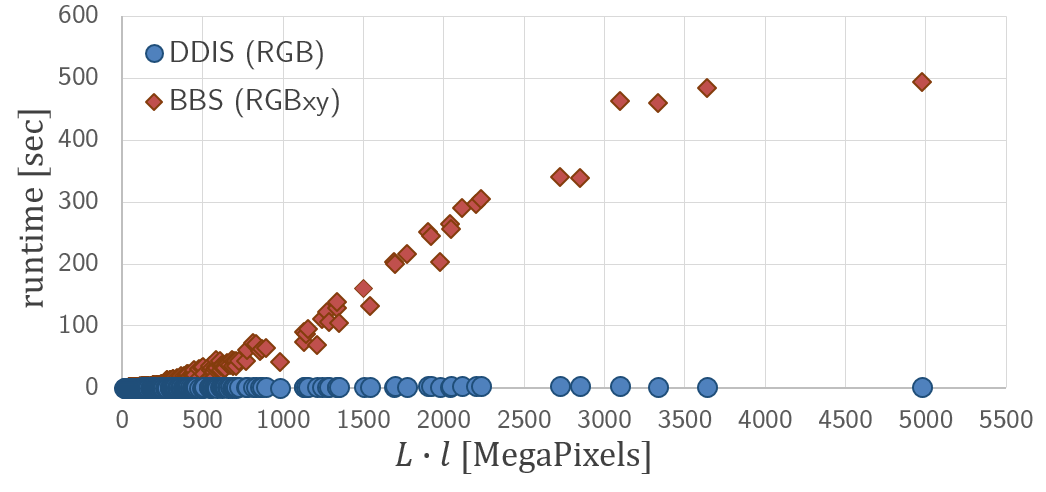}
		\end{subfigure}~
		\vspace{-0.5cm}
	\end{center}   
	\caption{\textbf{Runtime:} Each point in the scatter plot marks the runtime for one of the 270 image-pairs in the dataset ($dFrame\!=\!25$). For DDIS the runtime is always small while for BBS it becomes very long as the template and image size increase. 
	}
	\vspace{-0.5cm}
	\label{fig:runtime}
\end{figure}
 
\definecolor{bbs}{rgb}{0.772,	0.350,	0.066}
\definecolor{dis}{rgb}{0.517,	0.517,	1}
\definecolor{ddis}{rgb}{1,	0.102,	0.725}

\begin{figure*}
\centering
%		\rotatebox{90}{\hspace{0.05cm}}~
%		\vspace{0.075cm}
%		\begin{subfigure}[b]{.1925\textwidth}\includegraphics[width=.9975\linewidth]{TemplateMatching/runSamples/bike/T.png}\end{subfigure}~
%		\begin{subfigure}[b]{.1925\textwidth}\includegraphics[width=.9975\linewidth]{TemplateMatching/runSamples/bike/Target.png}\end{subfigure}~
%		\begin{subfigure}[b]{.1925\textwidth}\includegraphics[width=.9975\linewidth]{TemplateMatching/runSamples/bike/BBS_map.png}\end{subfigure}~
%		\begin{subfigure}[b]{.1925\textwidth}\includegraphics[width=.9975\linewidth]{TemplateMatching/runSamples/bike/DIS_map.png}\end{subfigure}~
%       \begin{subfigure}[b]{.1925\textwidth}\includegraphics[width=.9975\linewidth]{TemplateMatching/runSamples/bike/DDIS_map.png}\end{subfigure}
%        
   		\rotatebox{90}{\hspace{0.05cm}}~
   		\vspace{0.075cm}
   		\begin{subfigure}[b]{.1925\textwidth}\includegraphics[width=.9975\linewidth]{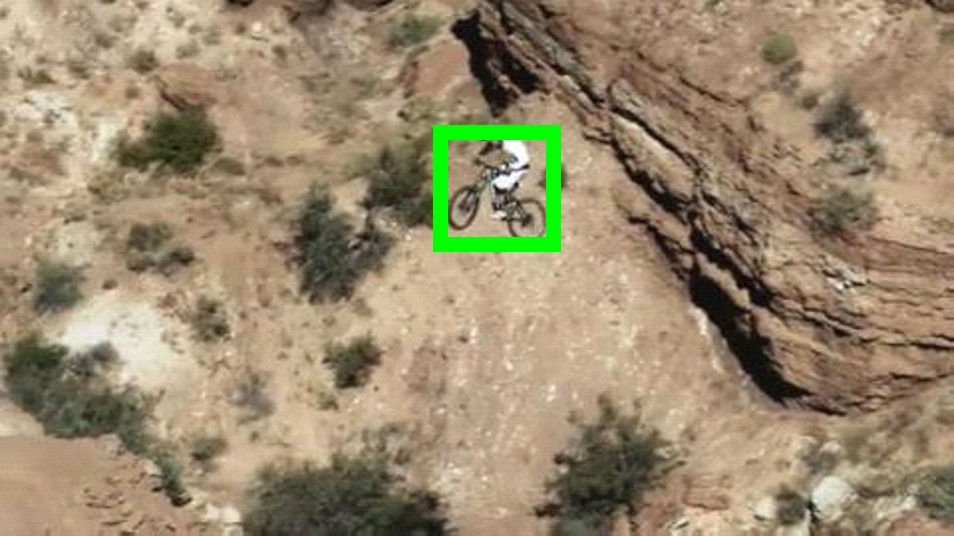}\end{subfigure}~
   		\begin{subfigure}[b]{.1925\textwidth}\includegraphics[width=.9975\linewidth]{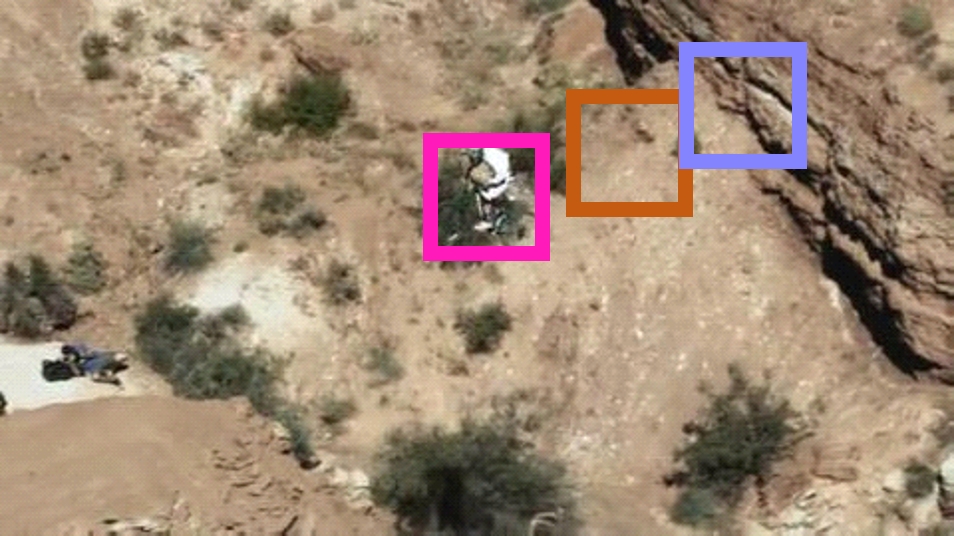}\end{subfigure}~
   		\begin{subfigure}[b]{.1925\textwidth}\includegraphics[width=.9975\linewidth]{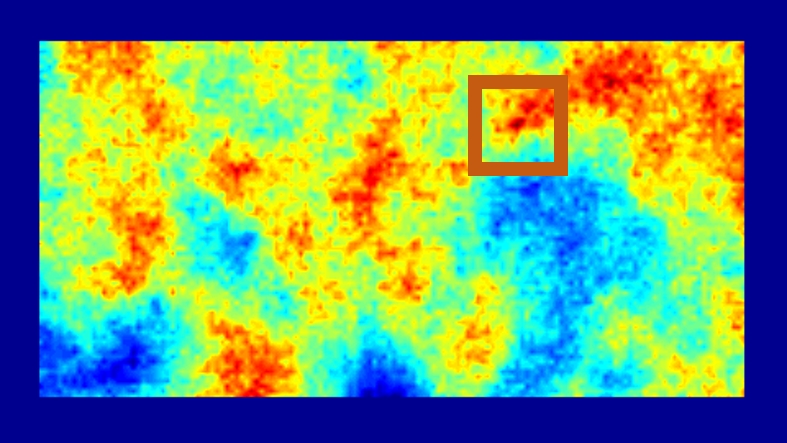}\end{subfigure}~
   		\begin{subfigure}[b]{.1925\textwidth}\includegraphics[width=.9975\linewidth]{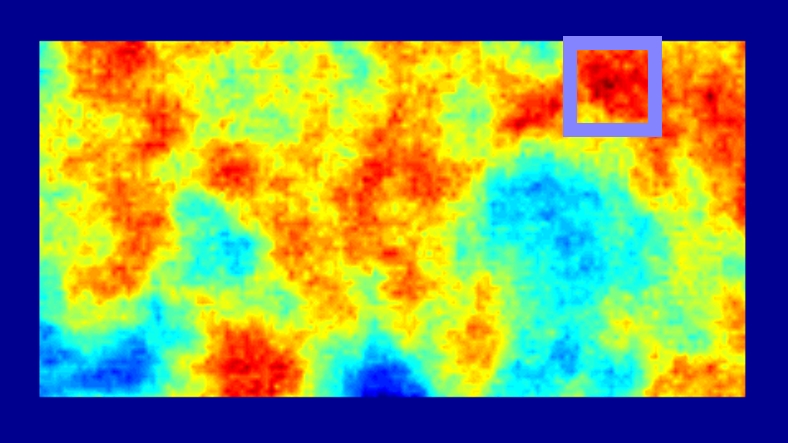}\end{subfigure}~
   		\begin{subfigure}[b]{.1925\textwidth}\includegraphics[width=.9975\linewidth]{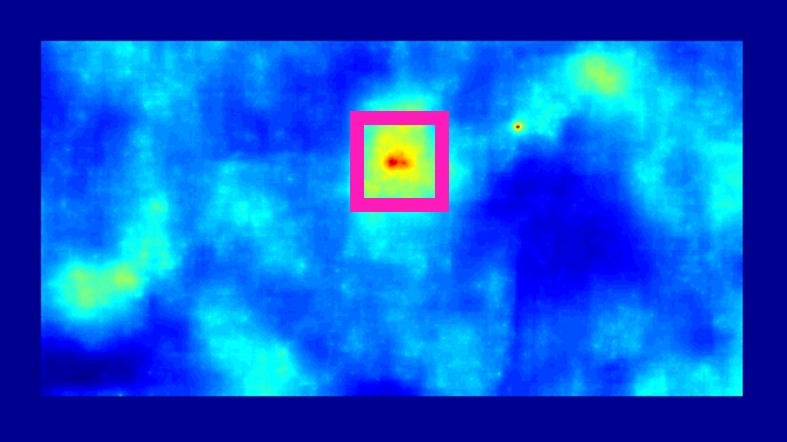}\end{subfigure}        		
		\rotatebox{90}{\hspace{0.05cm}}~
		\vspace{0.075cm}
		\begin{subfigure}[b]{.1925\textwidth}\includegraphics[width=.9975\linewidth]{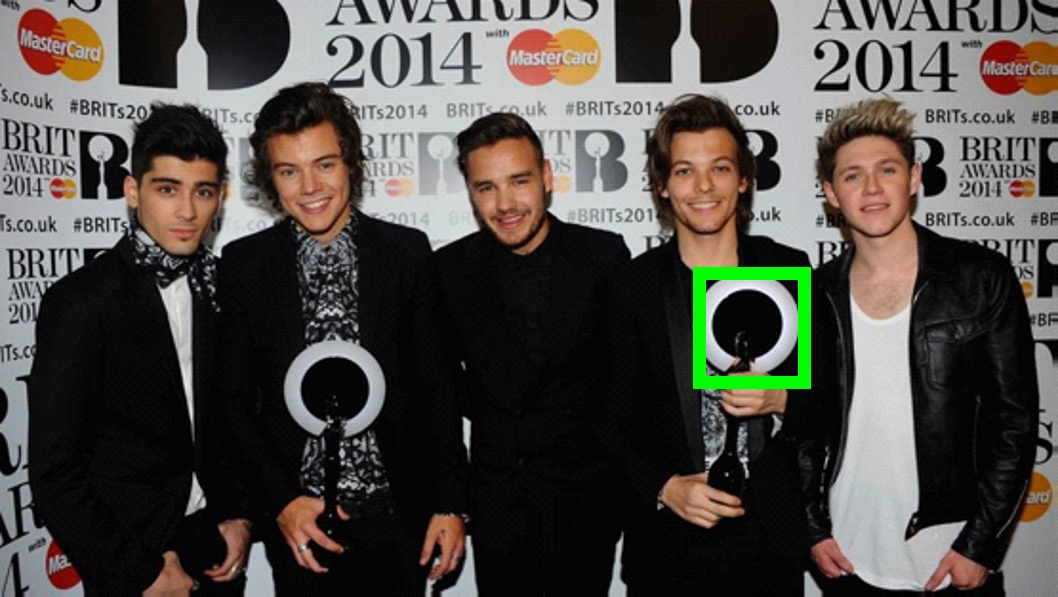}\end{subfigure}~
		\begin{subfigure}[b]{.1925\textwidth}\includegraphics[width=.9975\linewidth]{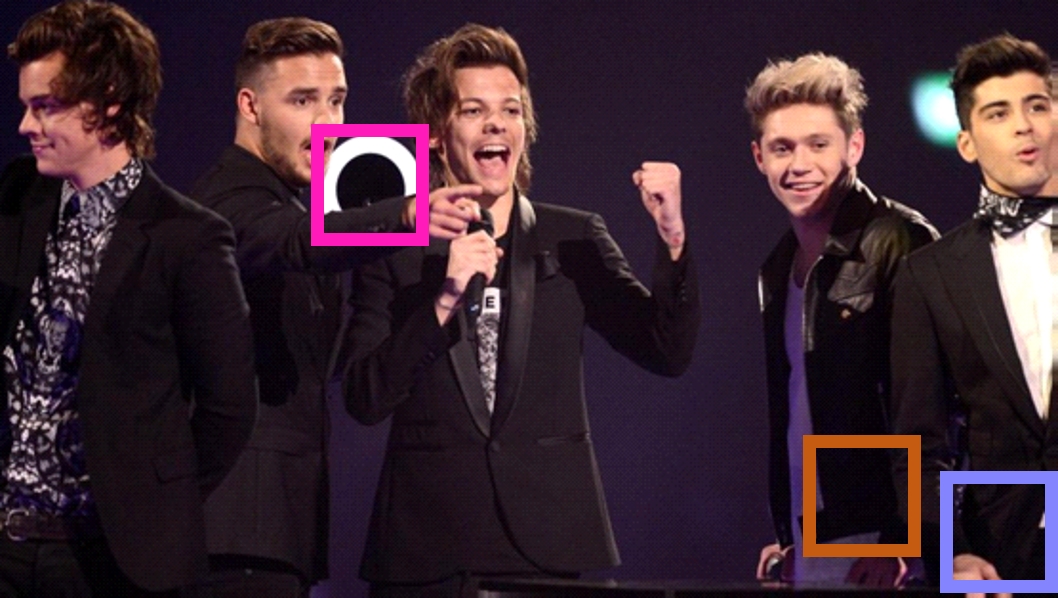}\end{subfigure}~
		\begin{subfigure}[b]{.1925\textwidth}\includegraphics[width=.9975\linewidth]{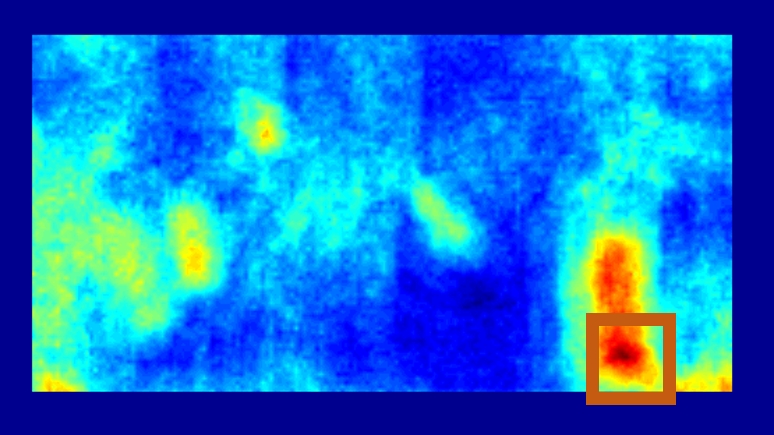}\end{subfigure}~
		\begin{subfigure}[b]{.1925\textwidth}\includegraphics[width=.9975\linewidth]{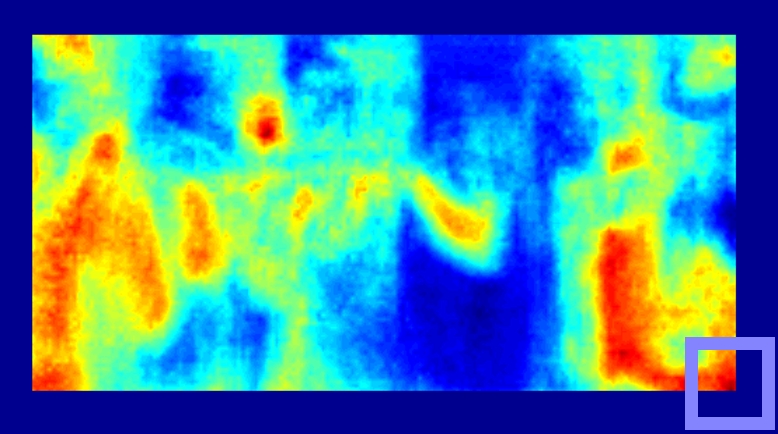}\end{subfigure}~
        \begin{subfigure}[b]{.1925\textwidth}\includegraphics[width=.9975\linewidth]{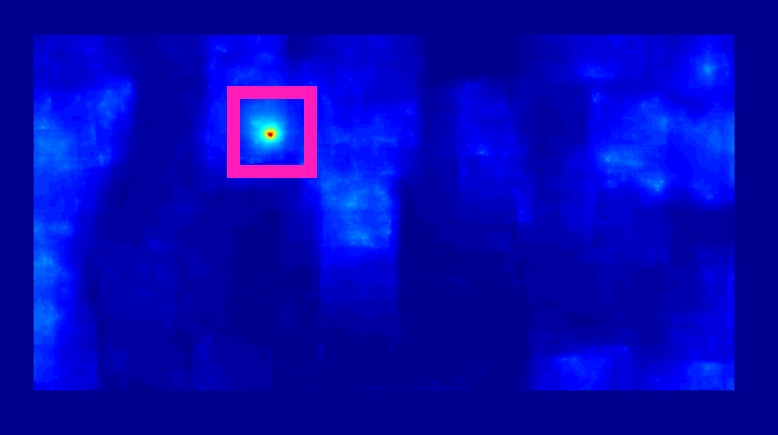}\end{subfigure}
		\rotatebox{90}{\hspace{0.05cm}}~
		\vspace{0.075cm}
		\begin{subfigure}[b]{.1925\textwidth}\includegraphics[width=.9975\linewidth]{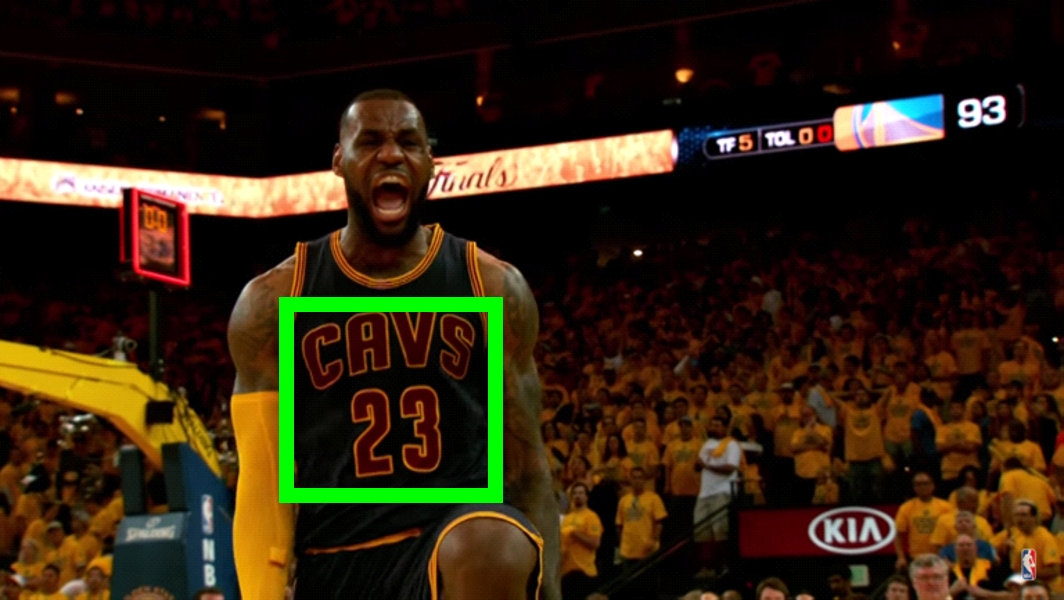}\end{subfigure}~
		\begin{subfigure}[b]{.1925\textwidth}\includegraphics[width=.9975\linewidth]{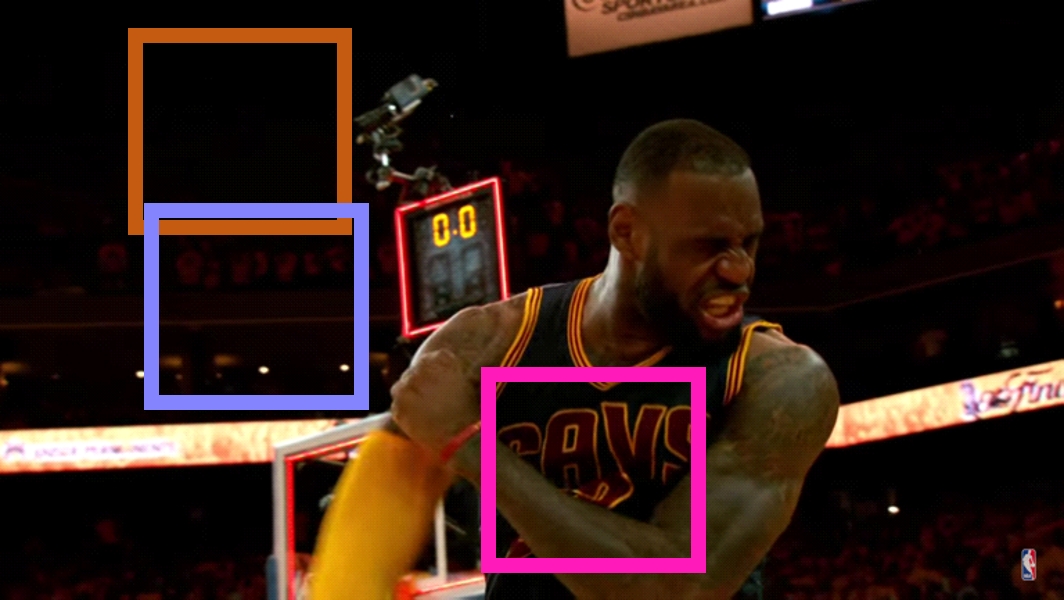}\end{subfigure}~
		\begin{subfigure}[b]{.1925\textwidth}\includegraphics[width=.9975\linewidth]{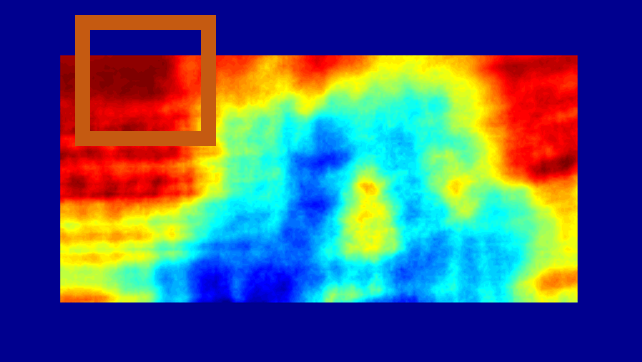}\end{subfigure}~
		\begin{subfigure}[b]{.1925\textwidth}\includegraphics[width=.9975\linewidth]{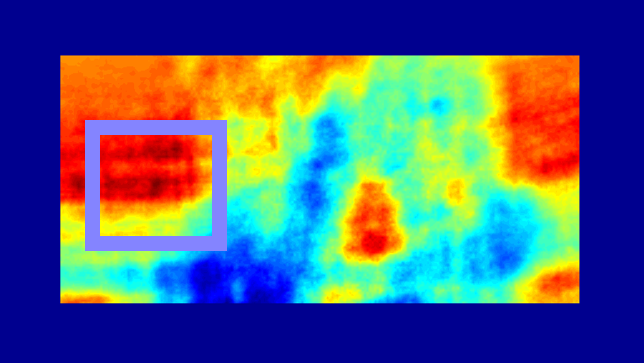}\end{subfigure}~
        \begin{subfigure}[b]{.1925\textwidth}\includegraphics[width=.9975\linewidth]{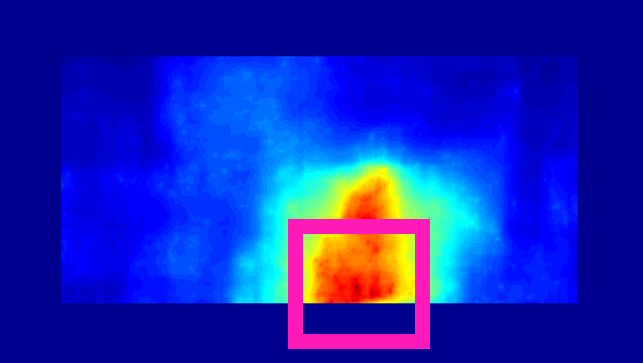}\end{subfigure}
	\caption{\textbf{Qualitative assessment}: The template marked in green (a), is detected in the target image (b) using three Template Matching methods: \textcolor{bbs}{\textbf{BBS}}, \textcolor{dis}{\textbf{DIS}} and \textcolor{ddis}{\textbf{DDIS}} (all using the RGB features). (c-e) The corresponding detection likelihood maps show that DDIS yields more peaked maps that more robustly identify the template. Going over the rows from top to bottom: (1) BBS prefers a target location where the background matches the template over the location where the motorcycle is at. This happens because the motorcycle deforms and hence there are few bi-directional correspondences between its template appearance and target appearance. DIS and DDIS use more information -- they consider all the one-directional correspondences. Therefore, they locate the motorcycle correctly. (2) The trophy won by the band \emph{One Direction} is fully seen in the template, but occluded in the target. Nonetheless, DDIS finds it (as Section~\ref{sec:youOnlyNeedOneDirection} said, we only need one direction...). (3) Complex deformations together with occlusion confuse both DIS and BBS, but not DDIS.}
    \vspace{-0.2cm}
	\label{fig:TM_examples}
\end{figure*}

\textbf{Runtime:} Our implementation is in MATLAB/c++ and all experiments were performed on a 32GB RAM, Intel i7 quad-core machine. The average(std) runtime for an image-pair in the benchmark, using color features is ~$0.86s (0.59)$, depending on the template size. For comparison, the average(std) time for BBS is orders of magnitude longer: $35.47s (80.36)$. The max and min runtimes of DDIS are $3.44s$ and $0.06s$, respectively, and for BBS are $493s$ and $0.14s$, respectively. 
Detailed results for $dFrame=25$ are presented in Figure~\ref{fig:runtime}.
This matches our complexity analysis that showed that DDIS is less affected by the template size, while BBS dependence on $l$ is polynomial.

\section{Conclusions}
\label{sec:conclusion}

We introduced a new approach for template matching in the wild, based on properties of the NN field of matches between target and template features. Our method suggests not only improvement in terms of detection accuracy, but also in terms of computational complexity.
A drawback of our algorithm is not dealing with significant scale change of the object. This could possibly be addressed, by computing the likelihood maps over multiple scales. A future research direction is to explore consideration of more than the first NN for each patch. This could be beneficial to handle repetitive textures.

An important observation, our analysis makes, is that one does not necessarily need bi-directional matches to compute similarity. This raises questions regarding the celebrated bi-directional-similarity approach, which provided excellent results, but was heavy to compute.

\paragraph{Acknowledgements}
This research was supported by the Israel Science Foundation Grant 1089/16 and by the Ollendorf foundation.

\begin{appendix}
\section{Appendix: DIS Expectation Term}
\label{appendix:DIS_Expectation}

In this appendix we develop mathematical expressions for the \textit{expectation} of DIS in $\mathbb{R}^1$.
We start by rewriting DIS in a form convenient for our derivations:
\begin{align*}
 & \underset{Q{\rightarrow}P}{\textbf{DIS}} = \frac{1}{N} \cdot \sum_{i=1}^{N}{dis_i(Q,P)} \\
 & dis_i(Q,P)=\mathbb{I}
	\left[ 
	\left\{q_j\!\in\! Q\! : \mathrm{NN}(q_j,P)\!=p_i\!\right\} \neq \emptyset
    \right]
\end{align*}
where $\mathbb{I}$ is an indicator function and $dis_i(Q,P)$ indicates whether $p_i$ is chosen as a NN match at least once. 

We proceed with the expectation:
\begin{equation}
E\left[ \underset{Q{\rightarrow}P}{\textbf{DIS}} \right] = 
\frac{1}{N}\!\sum_{i=1}^{N}{E\left[dis_i(Q,P)\right]} = E\left[dis_k(Q,P)\right]
\end{equation}
where the last step is since samples $Q$ and $P$ are drawn independently, so all indexes behave alike and we can choose some arbitrary index $k$. Continuing with the expectation of the indicator function, we have:
\begin{align}
{E\left[dis_k(Q,P)\right]} & =  {\mathrm{Pr}\left\{ dis_k(Q,P)=1 \right\}} \notag\\
& = { 1-\mathrm{Pr}\left\{ dis_k(Q,P)\!=\!0 \right\} }
\end{align}
\begin{claim}
\begin{align}
\label{eq:claim}
& {\mathrm{Pr}\left\{ dis_i(Q,P)\!=\!0 \right\}= }\\
& ={\int_{p_1}\!\cdots\!\int_{p_N} {(F_Q(p_{i}^{-})+1-F_Q(p_{i}^{+}))^N\!\cdot\!\prod\limits_{k=1}^{N} f_P(p_k)dp_k}} \notag
\end{align}
where $F_Q(x)\!=\!\mathrm{Pr}\left\{ q \leq x\right\}$ and $F_P(x)\!=\!\mathrm{Pr}\left\{p\leq x\right\}$ are the CDF's of Q and P, respectively. $p_{i}^{+},p_{i}^{-}$ are defined by:
\begin{align}
%\begin{equation}
\label{eq:pPlus}
p_{i}^{+} = p_i +
  \displaystyle\min_{\substack{p_k \in P\cup \left\{+\infty\right\} \\ p_k>p_i}} |p_k-p_i|/2\!\nonumber\\
%\end{equation}
%
%\begin{equation}\label{eq:pMinus}
p_{i}^{-} = p_i -
  \displaystyle\min_{\substack{p_k \in P\cup \left\{-\infty\right\} \\ p_k<p_i}} |p_k-p_i|/2\!
%\end{equation}
\end{align}
\end{claim}
\begin{claimproof}
Given a known set of samples P, the probability that the NN match for a sampled $q{\sim}Q$ is NOT $p_i$ is: 
\begin{align}
 \mathrm{Pr} & \left\{ \mathrm{NN}(q,P){\neq} p_i  \middle| P\right\}
   = \int_{-\infty}^{\infty} \mathbb{I}
	\left[ 
	\mathrm{NN}(q,P){\neq}p_i
    \right]f_Q(q)dq \notag\\
   & = 
     \int_{-\infty}^{p_{i}^{-}}f_Q(q)dq + \int_{p_{i}^{+}}^{\infty}f_Q(q)dq  \notag\\
    & = 
     F_Q(p_{i}^{-})+1-F_Q(p_{i}^{+})
     \label{eq:NniGivenP}
\end{align}
where we split $\mathbb{R}$ into two ranges where the indicator is not zero. 
%as illustrated in Figure~\ref{fig:qRangeIllustration}. 
Since $Q$ consists of $N$ independently sampled points, the probability that $p_i$ is not a NN match for any $q{\in}Q$ when $Q$ is sampled and $P$ is known, is:
\begin{align} 
\label{eq:disGivenP}
\mathrm{Pr} \left\{ dis_i(Q,P)=0 \middle| P\right\}  = 
\Big[ \mathrm{Pr}\left\{ \mathrm{NN}(q,P){\neq}p_i \middle| P\right\} \Big]^N 
\end{align}
Finally, since all of the points are sampled independently, we have:
\begin{align} 
\label{eq:almostProof}
& \mathrm{Pr}\left\{ dis_i(Q,P)\!=\!0 \right\}= \\
& = \int_{p_1}\!\cdots\!\int_{p_N} \mathrm{Pr} \left\{ dis_i(Q,P)=0 \middle| P\right\}\cdot\prod\limits_{k=1}^{N} f_P(p_k)dp_k \notag
\end{align}
Substituting equations~\eqref{eq:NniGivenP} and~\eqref{eq:disGivenP} in~\eqref{eq:almostProof} results in~\eqref{eq:claim}.
\end{claimproof}

%\begin{figure}
%    \begin{center}
%		\includegraphics[width=1.0\linewidth]{YouOnlyNeedOneDirection/ExpactationAnalysis/1dAnalysisIntuition.png}\vspace{-0.3cm}
%	\end{center}
%	\caption{Illustration of the range of possible values for a sampled point $q{\sim}Q$ such that $q$ is NOT a NN of some $p{\in}P$. In the current graph, $P\!=\!\left\{p_L, p_M, p_R\right\}$ consists two extremum points and one middle point, showing all the possible cases. The valid range is shown under the axis by three colors, orange, blue and green, for $p_L, p_M, p_R$ respectively.}
%    \vspace{-0.5cm}
%	\label{fig:qRangeIllustration}
%\end{figure}

\end{appendix}

%-----------------------------------------------------------------------------------
%\section*{Acknowledgements}
\clearpage
{\small
\bibliographystyle{ieee}
\bibliography{paper}
}

\end{document}